%% file: main.tex
\documentclass[a4paper,fleqn]{cas-dc}

\usepackage[authoryear]{natbib}
\usepackage{amsmath,amssymb,mathtools}
\usepackage{graphicx,booktabs,adjustbox,multirow,array}
\usepackage{pifont,makecell}
\usepackage{xcolor}
\usepackage{url}
\usepackage{xurl}

\usepackage{geometry}
\usepackage{titletoc}

\newcommand{\myparagraph}[1]{\par\addvspace{3pt}\noindent\textbf{#1.}\enspace\ignorespaces}

\begin{document}
\let\WriteBookmarks\relax
\def\floatpagepagefraction{1}
\def\textpagefraction{.001}

% Short title
\shorttitle{AgriChrono}

% Short author
\shortauthors{Jeong et al.}

% Main title of the paper
\title[mode=title]{AgriChrono: A Longitudinal Dataset and Benchmark for Novel-View Synthesis of Field-Grown Crops}

% Title footnote mark
% \tnotemark[1]

% Title footnote 1
% \tnotetext[1]{}

% =========================================================
% Authors
% =========================================================

% First author
\author[1,2]{Jaehwan Jeong}

% Footnote of the first author
\fnmark[1]

% Credit authorship
\credit{Conceptualization, Methodology, Software,
Resources, Investigation, Data curation,
Formal analysis, Visualization, Writing -- original draft}

% Second author
\author[2]{Tuan-Anh Vu}

% Credit authorship
\credit{Conceptualization, Writing -- review \& editing}

% Third author
\author[3]{Mohammad Jony}

% Credit authorship
\credit{Investigation}

% Fourth author
\author[3]{Shahab Ahmad}

% Credit authorship
\credit{Investigation}

% Fifth author
\author[3]{Md. Mukhlesur Rahman}

% Credit authorship
\credit{Resources, Validation}

% Sixth author
\author[1]{Sangpil Kim}

% Credit authorship
\credit{Supervision, Funding acquisition}

% Seventh author
\author[2]{M. Khalid Jawed}

% Corresponding author indication
\cormark[1]

% Email id of the seventh author
\ead{khalidjm@seas.ucla.edu}

% Credit authorship
\credit{Supervision, Project administration, Resources}

% =========================================================
% Affiliations
% =========================================================

% Affiliation 1: Korea University
\affiliation[1]{
    organization={Department of Artificial Intelligence, Korea University},
    city={Seoul},
    postcode={02841},
    country={Republic of Korea}
}

% Affiliation 2: UCLA
\affiliation[2]{
    organization={Department of Mechanical \& Aerospace Engineering, University of California, Los Angeles},
    city={Los Angeles},
    postcode={90095},
    state={CA},
    country={USA}
}

% Affiliation 3: North Dakota State University
\affiliation[3]{
    organization={Department of Plant Sciences, North Dakota State University},
    city={Fargo},
    postcode={58102},
    state={ND},
    country={USA}
}

% =========================================================
% Corresponding Author and Footnotes
% =========================================================

% Corresponding author text
\cortext[1]{Corresponding author}

% Footnote text
\fntext[1]{%
    \hangindent=2.3em
    \hangafter=1
    Jaehwan Jeong conducted this research while serving as a visiting researcher at the University of California, Los Angeles (UCLA).
}

% For a title note without a number/mark
% \nonumnote{}

% =========================================================
% Here goes the abstract

\begin{abstract}
    Reliable three-dimensional (3D) reconstruction of field-grown crops remains challenging due to thin and overlapping plant structures, wind-induced foliage motion, and variable natural illumination.
    Longitudinal agricultural datasets and standardized novel-view synthesis (NVS) benchmarks for field conditions remain limited.
    We present an integrated framework connecting robotic acquisition, dataset construction, NVS benchmarking, and agricultural scene reconstruction.
    We develop \textbf{AgriChrobot}, a robotic platform supporting web-based teleoperation and repeated multi-view acquisition along crop rows.
    Using this platform, we construct \textbf{AgriChrono}, a longitudinal canola dataset comprising 160 recording sessions over approximately one month, with four RGB streams, two depth maps, and camera-pose estimates.
    We establish seven NVS evaluation scenarios comprising four time-of-day and three acquisition-date scenarios, and evaluate 15 NeRF- and Gaussian Splatting-based methods.
    Beyond whole-image quality, we evaluate fine-detail reproduction within manually annotated canola regions using gradient- and edge-based metrics.
    The benchmark reveals that competitive whole-image performance does not necessarily translate into faithful reproduction of fine visual details in thin and overlapping crop structures.
    Motivated by this finding, we propose \textbf{AgriSplat}, which relaxes initial visibility filtering and reallocates capacity through contribution-based pruning and budgeted reinvestment into sub-threshold candidates ranked by existing gradient statistics.
    After anchor reallocation, asymmetric supervision penalizes insufficient contrast along reference image-gradient directions.
    Compared with Octree-GS, AgriSplat improves the crop-region gradient-energy ratio from 0.506 to 0.847 and edge F-score from 0.712 to 0.797, averaged across seven scenarios, while also improving whole-image PSNR, SSIM, and LPIPS.
    Together, these contributions support systematic research on scene-level crop reconstruction under field conditions.
\end{abstract}

% Use if graphical abstract is present
%\begin{graphicalabstract}
%\includegraphics{}
%\end{graphicalabstract}

% Highlights are provided separately in highlights.tex for journal submission.

%\nocite{*}

% Keywords
% Each keyword is seperated by \sep

\maketitle

% Main text
% =========================================================
% Main Text
% =========================================================

% --------------------- Main text -------------------------
\input{sections/01_introduction}
\input{sections/02_related_work}
\input{sections/03_agrichrobot}
\input{sections/04_agrichrono}
\input{sections/05_agrisplat}
\input{sections/06_experiments}
\input{sections/07_conclusion}

% ------------------ Submission statements ---------------
% Public access is verified; these pages also link to the Box-hosted files.
\section*{Data availability}
The AgriChrono repository (\url{https://github.com/StructuresComp/agri-chrono}) provides access to the AgriChrobot hardware designs and control software, the AgriChrono dataset, the NVS benchmark, and the AgriSplat implementation.
The repository also provides download instructions and dataset splits for reproducibility.

% =========================================================
% CRediT Authorship Contribution Statement
% =========================================================
% To print the credit authorship contribution details
\printcredits

% =========================================================
% Declaration of Competing Interest
% =========================================================

\section*{Declaration of competing interests}
The authors declare that they have no known competing financial interests or personal relationships that could have appeared to influence the work reported in this paper.

% =========================================================
% Acknowledgements
% =========================================================

\section*{Acknowledgments}

The authors thank Shivam Panda, Esther Gérard, Mason (Ziming) Zhao, Radha Lahoti, Yongkyu Lee, and Zhuonan Hao, current and former members of the Structures-Computer Interaction (SCI) Laboratory at the University of California, Los Angeles, for their helpful discussions and support during this research.
We also thank Gyusam Chang of Korea University for his discussions and assistance.

\section*{Funding}
This work was supported in part by the US Department of Agriculture (grant numbers 2024-67021-42528 and 2022-67022-37021), the Culture, Sports, and Tourism R\&D Program through the Korea Creative Content Agency grant funded by the Ministry of Culture, Sports and Tourism in 2024 (International Collaborative Research and Global Talent Development for the Development of Copyright Management and Protection Technologies for Generative AI, RS-2024-00345025), and the Institute of Information \& Communications Technology Planning \& Evaluation (IITP) grant funded by the Korea government (MSIT) (No. RS-2019-II190079, Artificial Intelligence Graduate School Program, Korea University).
This research was also supported by the AI Seoul Tech Research Support Program of the Seoul Future Foundation.

% \section*{Acknowledgment}
% This work was supported in part by the US Department of Agriculture (grant numbers 2024-67021-42528 and 2022-67022-37021), the Culture, Sports, and Tourism R\&D Program through the Korea Creative Content Agency grant funded by the Ministry of Culture, Sports and Tourism in 2024 (International Collaborative Research and Global Talent Development for the Development of Copyright Management and Protection Technologies for Generative AI, RS-2024-00345025), and the Institute of Information \& Communications Technology Planning \& Evaluation (IITP) grant funded by the Korea government (MSIT) (No. RS-2019-II190079, Artificial Intelligence Graduate School Program, Korea University).
% This research was also supported by the AI Seoul Tech Research Support Program of the Seoul Future Foundation.

% =========================================================
% References
% =========================================================
%% Loading bibliography style file
%\bibliographystyle{model1-num-names}
\bibliographystyle{cas-model2-names}
\bibliography{cas-refs}

\input{supplementary/main_supplementary}

\end{document}

%% file: sections/01_introduction.tex
\section{Introduction}
\label{sec:introduction}

Smart farming increasingly integrates advances in artificial intelligence and robotics to support agricultural applications.
Notably, agricultural digital twins (ADTs) are gaining significant attention \citep{subeesh2025digitaltwin,espejel2024comparative,malik2024digital}, with field-acquired imagery increasingly used to reconstruct agricultural environments as three-dimensional (3D) scenes \citep{peng2025uavo,zhang2025wheat3dgs,agrigaussian2026}.
Such 3D representations complement conventional two-dimensional (2D) crop imagery by enabling agricultural scenes to be visualized from viewpoints not directly captured by cameras \citep{li2025survey,akhtar2024unlocking,an2017quantifying}.

However, reconstructing field-grown crops remains challenging despite recent advances in Neural Radiance Fields (NeRFs) and 3D Gaussian Splatting (3DGS) \citep{barron2023zipnerf,kerbl20233d}.
Agricultural scenes contain thin and overlapping plant structures, complex occlusions, and wind-induced non-rigid motion.
Natural illumination varies throughout the day, while crop development changes scene appearance over time.
These challenges motivate systematic evaluation across acquisition dates and times of day, with crop-region analysis complementing whole-image metrics.
Yet longitudinal agricultural datasets and standardized NVS benchmarks supporting such evaluations remain limited.

\begin{figure}[pos=t]
    \centering
    \vspace{4pt}
    \includegraphics[width=\columnwidth]{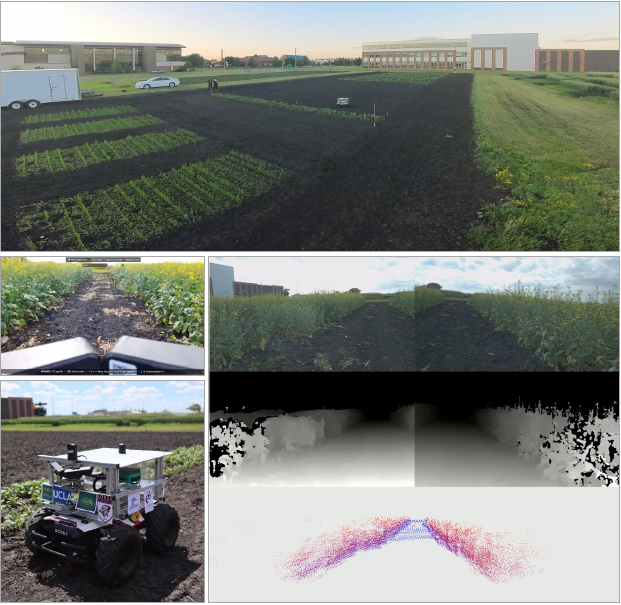}
    \caption{
        Overview of the AgriChrono field-scale data collection framework.
        \textbf{Top:} Agricultural field used for data acquisition.
        \textbf{Middle left:} Web-based interface for remote robot control and data collection.
        \textbf{Bottom left:} AgriChrobot equipped with multiple sensors.
        \textbf{Bottom right:} Representative RGB, depth, and auxiliary LiDAR observations.
    }
    \label{fig:study_overview}
\end{figure}

To support systematic evaluation and reconstruction, we develop \textbf{AgriChrobot}, a custom-designed robotic platform integrating multiple sensors with a web-based interface for remote operation and repeated data acquisition.
Repeated acquisition with this platform enables the construction of \textbf{AgriChrono}, a longitudinal canola dataset comprising 160 recording sessions collected over approximately one month.
These observations capture variations in crop appearance and natural illumination across acquisition dates and times of day, providing four RGB image streams, two stereo-derived depth maps, and corresponding visual--inertial odometry (VIO) camera-pose estimates.
Figure~\ref{fig:study_overview} illustrates the robotic platform and representative acquisition data.

Building upon AgriChrono, we establish a standardized NVS benchmark comprising seven evaluation scenarios, including four time-of-day and three acquisition-date scenarios.
Each scenario is reconstructed and evaluated separately to assess NVS performance across acquisition conditions.
We evaluate 15 NeRF- and 3DGS-based methods under a common protocol, considering whole-image rendering quality, fine-detail reproduction within manually annotated canola regions, and computational cost.
In addition to conventional whole-image metrics, we employ crop-region structural and perceptual similarity measures, together with gradient- and edge-based metrics.
The benchmark reveals that competitive whole-image performance does not necessarily translate into faithful reproduction of fine visual details in thin and overlapping crop structures.
Averaged across the seven scenarios, all 15 baseline methods exhibit crop-region gradient-energy ratios below one, indicating lower gradient energy in rendered canola regions than in the reference images.
Together with the edge-based evaluation, this observation highlights persistent limitations in fine-detail reproduction that may not be adequately reflected by whole-image metrics alone.

Motivated by these findings, we propose \textbf{AgriSplat}, an Octree-GS-based reconstruction method designed to improve fine-detail rendering in crop regions \citep{ren2024octree}.
AgriSplat combines Contribution-guided Anchor Reallocation (CAR) and Asymmetric Directional Supervision (ADS).
CAR relaxes initial visibility filtering to preserve fine-level anchor candidates and reallocates representation capacity through contribution-based pruning and budgeted reinvestment into sub-threshold candidates prioritized by existing gradient statistics.
After anchor growth and reallocation have ended, ADS penalizes insufficient rendered contrast along reference image-gradient directions using a one-sided constraint.
Together, CAR and ADS target fine-detail reproduction through representation allocation and subsequent image-space supervision.

Averaged across the seven scenarios, AgriSplat reduces whole-image LPIPS from 0.276 to 0.187 and crop-region LPIPS from 0.391 to 0.273 compared with Octree-GS.
It also improves the crop-region gradient-energy ratio from 0.506 to 0.847 and edge F-score from 0.712 to 0.797, alongside gains in whole-image PSNR and SSIM.
These results demonstrate improved perceptual rendering quality and fine-detail reproduction within crop regions under the evaluated field conditions.

The main contributions of this work are as follows:
\begin{enumerate}
    \item \textbf{AgriChrobot:}
    A custom-designed multi-sensor robotic platform supporting web-based remote operation and repeated multi-view field data acquisition.

    \item \textbf{AgriChrono:}
    A longitudinal canola dataset comprising 160 sessions capturing crop-development and illumination variations, providing RGB images, stereo-derived depth maps, and camera-pose estimates.

    \item \textbf{NVS Benchmark:}
    A standardized seven-scenario evaluation of 15 NeRF- and 3DGS-based methods, incorporating whole-image rendering quality, fine-detail reproduction within manually annotated canola regions, and computational cost.

    \item \textbf{AgriSplat:}
    A 3DGS-based reconstruction method combining CAR and ADS to improve rendering quality and fine-detail reproduction within canola regions.
\end{enumerate}

We make the AgriChrobot hardware designs and software, AgriChrono dataset, NVS benchmark, and AgriSplat implementation publicly available to support reproducible research and future developments in smart agriculture.

%% file: sections/02_related_work.tex
\section{Related Work}
\label{sec:related_work}

\subsection{Agricultural Datasets}
\label{subsec:related_datasets}

Agricultural datasets have been developed to support robotic perception, localization, crop monitoring, and plant phenotyping.
Rosario \citep{pire2019rosario} provides multi-sensor observations for localization and mapping in agricultural environments, while Under-Canopy \citep{cuaran2023undercanopy} provides robotic sensor data collected from corn and soybean fields at different growth stages.
Bacchus Long-Term \citep{polvara2024bacchus} supports long-term robotic localization and mapping through repeated vineyard observations under seasonal changes.
Other datasets focus on agricultural perception, including MinneApple \citep{hani2020minneapple} for fruit detection and segmentation, and CRDLD \citep{de2022deep} and AgroScapes \citep{panda2023agronav} for crop-row perception and robotic navigation.
GrowliFlower \citep{kierdorf2023growliflower} provides time-series UAV imagery for cauliflower growth analysis, whereas Crops3D \citep{zhu2024crops3d} and BonnBeetClouds3D \citep{marks2024bonnbeetclouds} provide point-cloud data for crop segmentation and phenotyping.
More recently, NIRPlant \citep{chang2026nirsplat} introduced multimodal plant observations under different lighting conditions for neural scene reconstruction.

Unlike agricultural datasets primarily designed for robotic localization, mapping, or point-cloud-based phenotyping, AgriChrono focuses on image-based novel-view synthesis under field conditions.
Its distinguishing feature is the combination of repeated multi-view acquisition across crop-development dates and times of day with a standardized NVS benchmark.
The seven scenarios enable systematic comparisons of reconstruction quality within individual acquisition sessions, incorporating both whole-image and crop-region evaluation.
Table~\ref{tab:related_dataset_tasks} compares the acquisition characteristics of related agricultural datasets.

\begin{table*}[pos=t]
\centering
\caption{
    Comparison of related agricultural datasets.
    \textbf{Growth Span} indicates repeated observations across major crop-development periods ($\checkmark$), partial coverage ($\triangle$), or no reported longitudinal acquisition ($\times$).
    \textbf{Illumination} indicates the number of explicitly defined lighting conditions or scheduled acquisition time slots; $\times$ denotes no reported systematic illumination sampling.
    \textbf{AgriChrono} comprises 160 traversal sessions across the primary observed growth period of canola, with four scheduled daily acquisition times.
}
\label{tab:related_dataset_tasks}

\renewcommand{\arraystretch}{1.1}
\setlength{\tabcolsep}{3pt}
\small
\adjustbox{max width=\textwidth}{
    \begin{tabular}{@{}lllcc@{}}
    \toprule
    \textbf{Dataset} & \textbf{Platform} & \textbf{Modalities} & \textbf{Growth Span} & \textbf{Illumination} \\
    \midrule
    Rosario \citep{pire2019rosario} & UGV & RGB, Depth, IMU, Pose & $\times$ & $\times$ \\
    MinneApple \citep{hani2020minneapple} & Handheld & RGB & $\times$ & 2 \\
    AgroScapes \citep{panda2023agronav} & UAV, UGV, Handheld & RGB & $\times$ & $\times$ \\
    GrowliFlower \citep{kierdorf2023growliflower} & UAV & RGB, Multispectral & $\checkmark$ & $\times$ \\
    Bacchus Long-Term \citep{polvara2024bacchus} & UGV & RGB, Depth, LiDAR, IMU, Pose & $\checkmark$ & $\times$ \\
    BonnBeetClouds3D \citep{marks2024bonnbeetclouds} & UAV & RGB-colored Point Clouds & $\times$ & $\times$ \\
    CRDLD \citep{de2022deep} & UGV & RGB & $\times$ & 4 \\
    Crops3D \citep{zhu2024crops3d} & Fixed, Handheld & RGB-colored Point Clouds & $\triangle$ & $\times$ \\
    Under-Canopy \citep{cuaran2023undercanopy} & UGV & RGB, Depth, IMU, Pose & $\checkmark$ & $\times$ \\
    NIRPlant \citep{chang2026nirsplat} & Fixed & RGB, Depth, NIR, LiDAR & $\times$ & 4 \\
    \midrule
    \textbf{AgriChrono (Ours)} & \textbf{UGV} & RGB, Depth, IMU, Pose & $\checkmark$ & \textbf{4} \\
    \bottomrule
    \end{tabular}
}
\end{table*}

\subsection{Neural Scene Reconstruction Methods}
\label{subsec:related_neural}

\myparagraph{Neural Radiance Fields}
Neural Radiance Fields (NeRFs) represent scene color and density using learned functions for novel-view synthesis.
Zip-NeRF \citep{barron2023zipnerf} combines grid-based features with anti-aliasing techniques, while Tetra-NeRF \citep{kulhanek2023tetranerf} employs a tetrahedral representation for radiance-field reconstruction.
SeaThru-NeRF \citep{levy2023seathru} incorporates scattering effects into the image formation model for underwater scene reconstruction.
Nerfacto \citep{nerfstudio}, provided within the Nerfstudio framework, combines hash encoding, proposal sampling, scene contraction, and per-image appearance modeling for neural rendering.

\myparagraph{3D Gaussian Splatting}
3D Gaussian Splatting (3DGS) represents scenes using anisotropic Gaussian primitives and visibility-aware rasterization \citep{kerbl20233d}.
For scene representation and optimization, Scaffold-GS \citep{scaffoldgs} organizes Gaussians around anchors with gradient-based growth and opacity-based pruning, while Octree-GS \citep{ren2024octree} and Hierarchical 3DGS \citep{hierarchicalgaussians24} employ hierarchical representations for different levels of detail.
3DGS-MCMC \citep{kheradmand20243d} reformulates Gaussian optimization and primitive management through a probabilistic interpretation, whereas Taming 3DGS \citep{Taming3DGS} focuses on reconstruction under limited computational resources.
gsplat \citep{ye2025gsplat} provides optimized implementations of Gaussian Splatting operations.

Recent studies also investigate Gaussian allocation, pruning, and fine-detail reconstruction.
AbsGS \citep{ye2024absgs} addresses gradient cancellation during densification by accumulating absolute positional gradients, whereas EAGLES \citep{girish2024eagles} prunes Gaussians based on accumulated opacity--transmittance contributions.
FreGS \citep{zhang2024fregs} employs progressive frequency-domain regularization, while HF-GS \citep{li2026hf} combines edge-weighted supervision, scale-aware refinement, and periodic pruning.

Other approaches address rendering, appearance modeling, and geometric reconstruction.
3D-GRT \citep{3dgrt2024} introduces Gaussian-based ray tracing, while 3DGUT \citep{wu20253dgut} extends Gaussian rendering to nonlinear camera projections and secondary rays.
GS-W \citep{zhang2024GS-W} addresses appearance variations and transient scene content in unconstrained image collections.
PGSR \citep{chen2024pgsr} incorporates geometric constraints for surface reconstruction.

\myparagraph{Agricultural 3D Gaussian Splatting}
Agricultural applications of 3DGS have addressed plant reconstruction and phenotyping across acquisition environments.
PlantGaussian \citep{shen2025plantgaussian} and P3DFusion \citep{wang2025p3dfusion} combine Gaussian reconstruction with plant segmentation and mesh extraction, while Cotton3DGaussians \citep{jiang2025cotton3dgaussians} focuses on individual cotton plants and architectural trait estimation.
For outdoor environments, 3DGS-Ag \citep{chen2025gsag} addresses multi-scale peach-orchard reconstruction, AgriGaussian \citep{agrigaussian2026} integrates depth supervision and appearance modeling for field-grown chili peppers, and Wheat3DGS \citep{zhang2025wheat3dgs} combines wheat-head reconstruction with instance segmentation and phenotyping.
Other studies emphasize geometric reconstruction.
Thin Gaussian Splatting \citep{shen2026thingaussian} employs flattened Gaussian primitives and geometric consistency constraints for peach-tree surface reconstruction, while GC-3DGS-Pheno \citep{wu2026gc3dgspheno} integrates depth, normal, and multi-view constraints for plant phenotyping.

Within 3DGS research, AgriSplat targets fine-detail reconstruction of field-grown crops.
Extending Octree-GS, CAR reallocates anchor capacity through contribution-based pruning and reinvestment into previously unselected, gradient-ranked candidates, differing from pruning-focused EAGLES and gradient-driven AbsGS.
Unlike edge-weighted HF-GS, ADS selectively penalizes insufficient contrast along reference-gradient directions after anchor growth, under a fixed anchor topology.

%% file: sections/03_agrichrobot.tex
\section{AgriChrobot: Data Collection Platform}
\label{sec:agrichrobot}

\begin{figure}[pos=h]
    \centering
    \vspace{3pt}
    \includegraphics[width=\columnwidth]{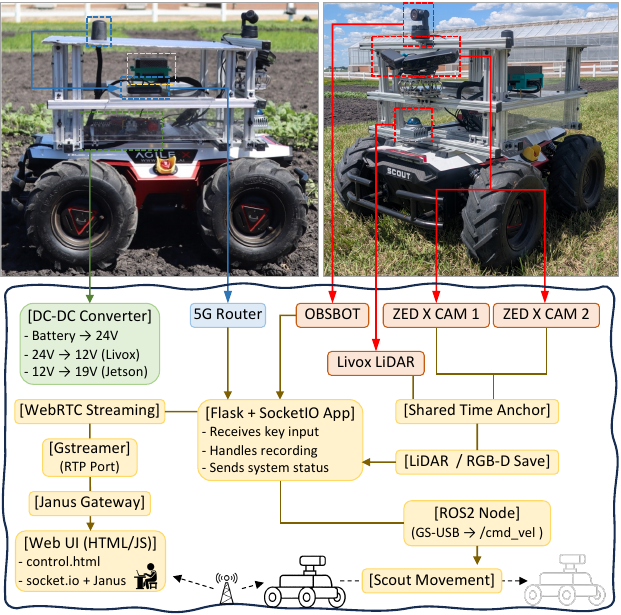}
    \caption{
        Hardware and software architecture of AgriChrobot.
        \textbf{Top:} The robotic platform with a custom-designed sensor assembly on an AgileX Scout 2.0 UGV.
        \textbf{Bottom:} Hardware arrangement and connections among sensing, onboard computing, power conversion, remote operation, and data recording modules.
        The software integrates WebRTC-based video streaming, Flask and Socket.IO-based control, session recording, and ROS~2-based vehicle control.
    }
    \label{fig:agrichrobot_hardware}
\end{figure}

\subsection{Design Objectives and Robotic Architecture}
\label{subsec:robot_objectives}

AgriChrobot was developed to support repeated ground-level observations of field-grown crops for constructing the AgriChrono dataset. The platform was designed to acquire multiple camera perspectives along crop rows and enable remote vehicle operation and data recording during repeated field visits. Built on an AgileX Scout 2.0 unmanned ground vehicle (UGV), AgriChrobot incorporates a custom sensor frame and protective chassis designed, fabricated, and assembled to accommodate imaging sensors, onboard computing, power conversion, and communication equipment.

The robotic assembly consists of three tiers. The lower tier accommodates power conversion modules and an auxiliary Livox Mid-360 LiDAR. The middle tier houses two ZED X stereo cameras, an NVIDIA Jetson AGX Orin (64~GB), and a 5G communication module. The upper tier supports an OBSBOT pan--tilt--zoom (PTZ) camera and an omnidirectional antenna for remote observation and operation. The sensor assembly also incorporates protective aluminum--acrylic panels and a sun shield for outdoor deployment. Figure~\ref{fig:agrichrobot_hardware} illustrates the hardware configuration and functional architecture of AgriChrobot. Detailed hardware specifications and mechanical design information are provided in Supplementary Material~S1.1.

\subsection{Stereo-Camera Layout}
\label{subsec:robot_cameras}

Two ZED X stereo cameras with 4-mm lenses are mounted on adjustable three-axis supports with passive vibration isolation. The cameras are oriented toward different forward-oblique directions to acquire views of crop structures during vehicle traversal. Each stereo unit provides left and right RGB images, yielding four RGB image streams in total. The camera arrangement enables successive observations of crop rows from different viewpoints, including surrounding soil and vegetation.

For each stereo unit, depth maps corresponding to the left camera images are generated using the ZED SDK, as described in Section~\ref{subsec:dataset_statistics}. Each unit also provides session-local visual--inertial odometry (VIO) camera-pose estimates. The stereo units use their respective intrinsic and within-unit extrinsic calibration parameters. A relative extrinsic transformation between the two stereo units was also estimated using a multi-sensor calibration procedure. The calibration parameters and estimated inter-unit transformation are provided in Supplementary Material~S1.2.

\subsection{Embedded Software and Teleoperation}
\label{subsec:robot_software}

The onboard software integrates ROS~2 Humble for vehicle control, a CAN-bus interface to the Scout platform, and a browser-based interface for remote operation.
A Flask and Socket.IO backend manages operator commands, recording control, and system-status information.
Video feedback is transmitted via WebRTC using GStreamer and Janus Gateway.
The interface supports vehicle movement, PTZ camera control, system monitoring, and recording management.
A zero-velocity stop command is issued when the browser connection is lost or the control interface loses focus.

During field acquisition, the two ZED X stereo units record camera streams in \texttt{.svo2} format. Recording processes are organized using a shared session-start time reference. Subsequent offline processing extracts RGB images, stereo-derived depth maps, and frame-indexed metadata, including VIO-estimated camera poses, to construct the AgriChrono dataset. This workflow separates field recording from the preparation of data products used in reconstruction experiments. Further details on the software environment, acquisition modules, and recording procedures are provided in Supplementary Material~S1.3.

\subsection{Field Deployment and Acquisition Procedure}
\label{subsec:robot_deployment}

AgriChrobot was deployed for repeated observations of a canola field at Site~1. The robot traversed the plot in clockwise (CW) and counterclockwise (CCW) directions, allowing the crop rows to be observed from different sides. The target traversal speed was approximately 0.2~m/s, and the stereo cameras recorded at approximately 15 frames/s. Repeated acquisitions were scheduled at different times of day throughout the observed crop-development period, with actual recording activities depending on field and weather conditions. A total of 160 traversal sessions were collected at Site~1 using the platform.

\begin{figure*}[pos=t]
    \centering
    \includegraphics[width=\textwidth]{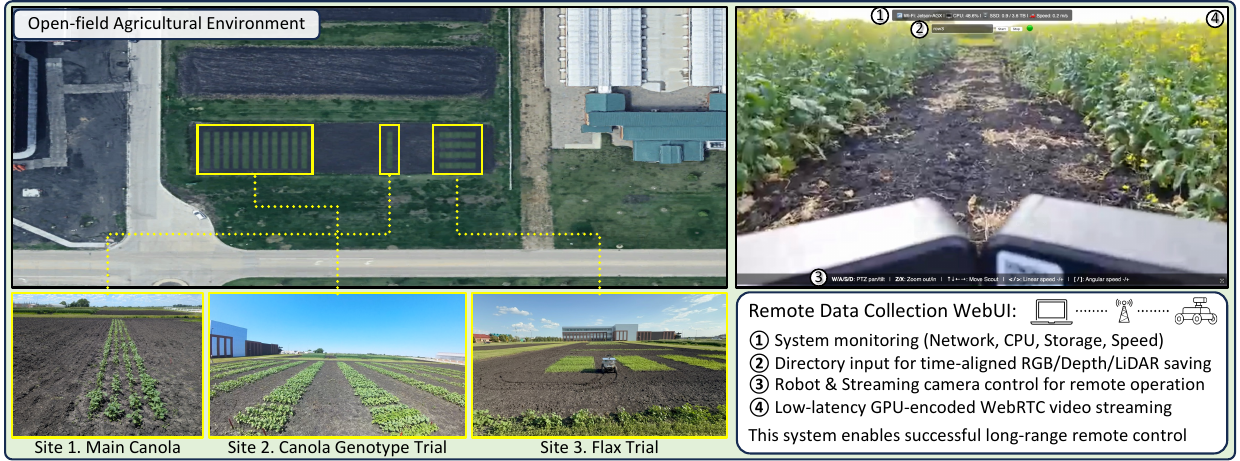}
    \caption{
        Field acquisition sites and web-based remote operation interface.
        \textbf{Site~1:} Main canola field used for repeated observations of crop development and natural illumination conditions.
        \textbf{Site~2:} Canola genotype trial containing multiple canola varieties.
        \textbf{Site~3:} Flax trial comprising plots with different cultivation treatments.
        \textbf{Web interface:} System monitoring (1), recording directory and session management (2), robot and PTZ camera control (3), and real-time video streaming through WebRTC (4). Site~1 provides the primary observations analyzed in this study; Sites~2 and 3 are documented in Supplementary Material~S2.
    }
    \label{fig:field_site_webui}
\end{figure*}

Figure~\ref{fig:field_site_webui} presents the field acquisition sites and the remote operating interface. In addition to Site~1, AgriChrobot was deployed at a canola genotype trial (Site~2) and a flax trial (Site~3). Although the dataset analysis and novel-view synthesis benchmark in this study focus on Site~1, recordings from Sites~2 and 3 are included in the broader acquisition archive. 
The AgriChrono dataset is described in Section~\ref{sec:agrichrono}, and additional information on Sites~2 and 3 is provided in Supplementary Material~S2.2.

%% file: sections/04_agrichrono.tex
\section{AgriChrono: Longitudinal Canola Dataset}
\label{sec:agrichrono}

\subsection{Field Site and Longitudinal Acquisition}
\label{subsec:dataset_scope}

AgriChrono was collected at the North Dakota State University experimental field in Fargo, North Dakota, USA, from July~2 to August~1, 2025, as shown in Figure~\ref{fig:field_site_webui}. The primary dataset analyzed in this study comprises observations of canola (\textit{InVigor L340PC}) at Site~1, a plot measuring approximately $15.2 \times 0.91$~m. The period spans approximately one month, covering changes in crop appearance and canopy structure during the observed growth period.

\begin{figure*}[pos=t]
    \centering
    \includegraphics[width=\textwidth]{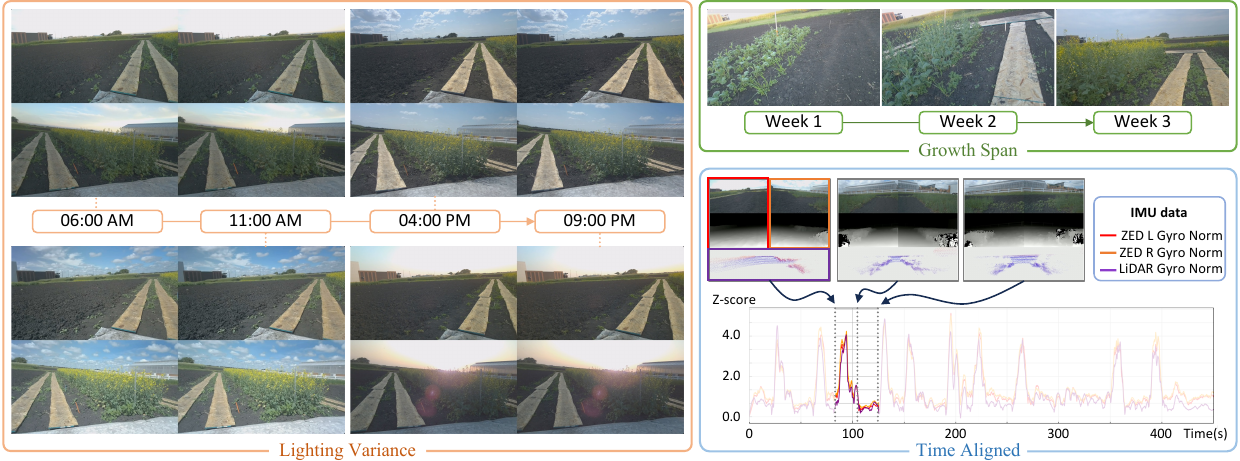}
    \caption{
        Visualization of dataset properties.
        \textbf{Lighting Variance:} RGB frames of the same scene at four times of day, showing natural illumination changes. The top two images are from the left stereo camera, and the bottom two are from the right stereo camera.
        \textbf{Growth Span:} Weekly progression (Week 1--3) of the same crop row, highlighting canola growth and structural change.
        \textbf{Time Alignment:} Z-score-normalized gyroscope magnitudes (L2 norm) from the two ZED cameras and LiDAR, illustrating temporal correspondence among the recorded sensor signals.
    }
    \label{fig:site1_illumination}
\end{figure*}

As described in Section~\ref{subsec:robot_deployment}, AgriChrobot traversed Site~1 in clockwise (CW) and counterclockwise (CCW) directions to capture the crop rows from different viewing directions. From July~2 to 21, acquisitions were scheduled at approximately 06:00, 11:00, 16:00, and 21:00 to observe the field under different natural illumination conditions. Figure~\ref{fig:site1_illumination} illustrates the resulting variations in illumination throughout the day and changes in canopy density and flowering appearance over successive weeks. After July~21, the acquisition frequency was reduced as crop growth slowed, while weather and field accessibility, including rain and muddy ground conditions, also affected the recording schedule. The detailed collection calendar is provided in Supplementary Material~S2.1, while additional recordings from a canola genotype trial (Site~2) and a flax trial (Site~3) are described in Supplementary Material~S2.2.

\subsection{Dataset Composition and Statistics}
\label{subsec:dataset_statistics}

The Site~1 dataset comprises 160 recording sessions obtained from 80 collection rounds, each consisting of one clockwise (CW) and one counterclockwise (CCW) traversal. These sessions contain 559,575 time-indexed samples recorded over 37,932~s (approximately 10.54~h), with an average frame rate of 14.7~FPS and a total data volume of approximately 13.7~TB. Table~\ref{tab:agri_site1_stats} summarizes the session counts, recording durations, sample counts, and data volumes for each traversal direction. The complete AgriChrono archive comprises 175 sessions across three field sites, with detailed statistics provided in Supplementary Material~S2.2.

\begin{table}[pos=t]
    \centering
    \caption{
        Statistics of the \textbf{AgriChrono} dataset at Site~1, detailing recording sessions, total duration, average frame rate, number of multi-modal samples, and overall data volume.
    }
    \renewcommand{\arraystretch}{1.1}
    \setlength{\tabcolsep}{3pt}
    \small
    \adjustbox{max width=\columnwidth}{
        \begin{tabular}{lccccc}
            \toprule
            \textbf{Site} & \textbf{\# Sessions} & \textbf{FPS} & \textbf{Duration (s)} & \textbf{\# Samples} & \textbf{Size} \\
            \midrule
            Site~1$_{\texttt{CW}}$ & 80 & 14.7 & 18,834 & 276,656 & 6.8\,TB \\
            Site~1$_{\texttt{CCW}}$ & 80 & 14.7 & 19,098 & 282,919 & 6.9\,TB \\
            \midrule
            \textbf{Total} & \textbf{160} & \textbf{14.7} & \textbf{37,932} & \textbf{559,575} & \textbf{13.7\,TB} \\
            \bottomrule
        \end{tabular}
    }
    \label{tab:agri_site1_stats}
\end{table}

\myparagraph{Dataset Properties}
Each recording session provides four $1920 \times 1080$ RGB image streams from two ZED X stereo cameras, together with two depth maps aligned with the respective left reference cameras. The depth maps are generated using the ZED SDK neural depth estimation mode, while each stereo unit independently estimates six-degree-of-freedom camera poses through visual--inertial odometry (VIO). These RGB, depth, and pose observations provide multiple viewpoints and corresponding geometric information along the crop rows within each recording session. Inter-frame displacement statistics for the camera-pose trajectories are provided in Supplementary Material~S2.3.

Across the 160 Site~1 sessions, AgriChrono extends these multi-view observations into a longitudinal record of the same agricultural field throughout the observed growth period. The dataset captures changes in canopy density, leaf and stem arrangements, and flowering structures as canola develops, alongside variations in natural illumination and image appearance at different times of day. The recorded scenes also contain thin and overlapping plant structures, partial occlusions, and surrounding soil and vegetation. By combining repeated observations of crop development with multi-view RGB images, stereo-derived depth, and VIO poses, AgriChrono provides a dataset for investigating image-based 3D reconstruction under varying crop-growth and illumination conditions.

\subsection{Data Organization and Public Availability}
\label{subsec:dataset_format}

\begin{figure}[pos=t]
    \centering
    \vspace{3pt}
    \includegraphics[width=\columnwidth]{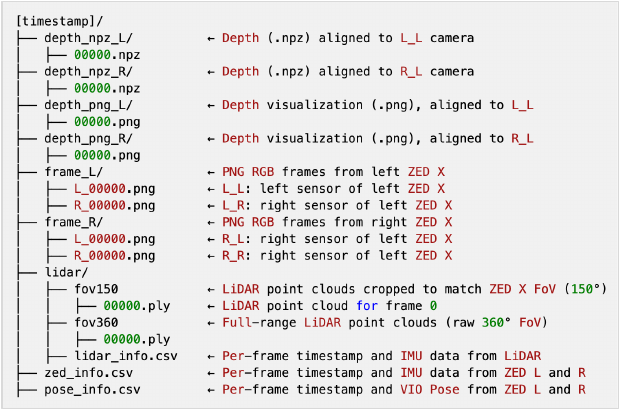}
    \caption{
        \textbf{Data structure of AgriChrono.}
        Each recording session contains four RGB image streams, two depth maps, and timestamp-indexed metadata for IMU measurements and 6-DoF VIO poses. The full acquisition archive also includes auxiliary LiDAR point clouds stored with two fields of view.
    }
    \label{fig:data_structure}
    \vspace{-5pt}
\end{figure}

As illustrated in Figure~\ref{fig:data_structure}, AgriChrono is organized by recording session, with raw sensor recordings and processed data products. The raw stereo-camera streams are preserved in \texttt{.svo2} format, while processed RGB data consist of rectified \texttt{.png} image pairs from the two stereo cameras. Depth maps are provided as numerical arrays in \texttt{.npz} format and visualization images in \texttt{.png} format. Frame-indexed metadata, including camera timestamps, IMU measurements, and six-degree-of-freedom VIO poses, are stored in \texttt{.csv} files such as \texttt{zed\_info.csv} and \texttt{pose\_info.csv}. This session-based organization preserves the association between recorded images, depth information, and camera trajectories across different acquisition dates and times. The full acquisition archive also includes LiDAR recordings.

For accessibility, the AgriChrono repository is released in four configurations: (i) the full 18\,TB multi-modal repository, (ii) a 1.2\,TB sample subset, (iii) a 162\,GB benchmark subset for NVS evaluation across seven scenarios, and (iv) a 5.4\,TB stereo-only RGB variant. The full repository provides the complete field acquisition records, while the smaller subsets offer access to selected observations and data prepared for reconstruction experiments. 

% Together, these releases support data exploration, method development, and reproducible evaluation of image-based 3D reconstruction methods.
% The dataset and download instructions are available through the \href{https://github.com/StructuresComp/agri-chrono/tree/main/datasets}{AgriChrono dataset repository}, while the associated hardware documentation and software are provided in the \href{https://github.com/StructuresComp/agri-chrono}{project repository}.

%% file: sections/05_agrisplat.tex
\clearpage
\section{AgriSplat: Agricultural Gaussian Splatting}
\label{sec:agrisplat}

\begin{figure*}[pos=t]
    \centering
    \includegraphics[width=\textwidth]{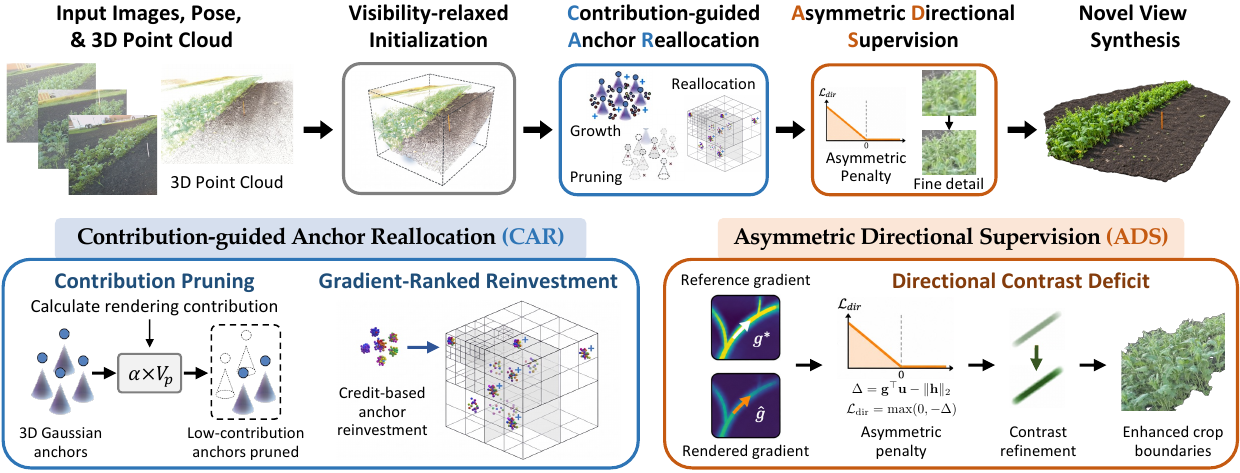}
    \caption{\textbf{Overview of AgriSplat.}
        Given input images and an initial 3D point cloud, AgriSplat begins with visibility-relaxed anchor initialization.
        During anchor growth, Contribution-guided Anchor Reallocation (CAR) prunes low-contribution anchors and reinvests the resulting credit into eligible sub-threshold candidates ranked by gradient magnitude.
        After anchor growth and reallocation, Asymmetric Directional Supervision (ADS) refines rendering under a fixed anchor topology while continuing to optimize the trainable Gaussian parameters.
        ADS penalizes insufficient rendered contrast along reference image-gradient directions to improve the reproduction of fine crop structures.
        The lower panels illustrate the CAR and ADS mechanisms.
    }
    \label{fig:agrisplat_overview}
\end{figure*}

AgriSplat builds upon Octree-GS to improve fine-detail rendering in agricultural scenes containing thin and overlapping plant structures.
The method combines Contribution-guided Anchor Reallocation (CAR), which preserves initially visible anchor candidates and reallocates representation capacity through contribution-based pruning and budgeted reinvestment, with Asymmetric Directional Supervision (ADS), which selectively penalizes insufficient local contrast after anchor allocation.

\subsection{Preliminaries}
\label{sec:agrisplat_prelim}

\myparagraph{Octree-GS}
Octree-GS~\citep{ren2024octree} constructs a hierarchical Gaussian representation by organizing an initial point cloud into a multi-resolution octree.
Each octree level contains Gaussian anchors that generate neural Gaussians through learned features and offsets, while a viewpoint-dependent level-of-detail (LOD) mechanism selects anchors for rendering.
At initialization, camera-to-anchor distance determines the eligible LOD for each training view, and anchor candidates are filtered according to their LOD selection frequency across training cameras using an adaptive visibility threshold.
As optimization proceeds, anchors are added based on accumulated offset-gradient statistics and removed according to opacity statistics.

\subsection{Contribution-guided Anchor Reallocation}
\label{sec:agrisplat_car}

During sequential crop-row traversals, each scene location is observed at close range by only a subset of the training cameras.
Since finer LODs in Octree-GS are primarily selected from nearby viewpoints, useful close-range observations may still yield low visibility ratios across the full set of training cameras.
Consequently, fine-level anchor candidates may be removed during visibility-based initialization despite having useful nearby observations.
To preserve these candidates, we relax the initialization filtering to retain anchors selected by at least one training camera.

Retaining a broader set of anchors, however, may preserve representations that contribute little to the rendered images.
CAR addresses this trade-off by coupling contribution-based pruning with budgeted reinvestment, reallocating capacity from weakly contributing anchors to eligible locations not selected by the original gradient-based growth rule.

\myparagraph{Contribution Pruning}
We measure the rendering contribution of each anchor by summing the alpha-compositing weights of its generated Gaussians across the training views.
For an anchor $a$, the contribution is defined as
\begin{equation}
    \qquad \quad W_a =
    \sum_{v \in \mathcal{V}_{\mathrm{train}}}
    \sum_{i \in \mathcal{G}_a^{(v)}}
    \sum_{p \in \Omega_v}
    \alpha_{i,p}^{(v)} T_{i,p}^{(v)}
    \label{eq:agrisplat_contribution}
\end{equation}
where $v$ indexes a training camera view, $\mathcal{V}_{\mathrm{train}}$ is the set of training views, $\mathcal{G}_a^{(v)}$ denotes the Gaussians generated by anchor $a$ in view $v$, and $\Omega_v$ is the corresponding image domain.
Here, $\alpha_{i,p}^{(v)}$ represents the opacity contribution of Gaussian $i$ at pixel $p$, including its projected spatial weighting, while $T_{i,p}^{(v)}$ denotes the preceding transmittance.
Their product is the Gaussian's alpha-compositing weight at the pixel. Summing these weights across Gaussians, pixels, and training views yields the anchor-level contribution $W_a$.

At each contribution-pruning round, $W_a$ is computed anew by evaluating the current Gaussian representation across all training views under their corresponding active LOD configurations.
Anchors with $W_a<\theta_W$ are removed, where $\theta_W$ is a shared contribution threshold across octree levels.
Anchors at levels not yet activated during progressive training are excluded to avoid removing candidates that have not been rendered.
This criterion supplements the original opacity-based pruning during optimization, while the number of anchors removed by contribution pruning is accumulated as reinvestment credit when subsequent growth rounds remain available.

\myparagraph{Gradient-Ranked Reinvestment}
The resulting credit is reinvested into locations proposed by offsets not selected by the original gradient-based growth rule.
Rather than introducing an additional gradient estimator, CAR reuses the view-space gradient statistics already accumulated by Octree-GS.
For an offset $j$, the statistic is defined as
\begin{equation}
    \quad g_j =
    \frac{1}{n_j}
    \sum_{t\in\mathcal{T}_j}
    \left\|
        \frac{\partial\mathcal{L}_t}
        {\partial\boldsymbol{\mu}_{j,t}^{2D}}
    \right\|_2,
    \qquad
    n_j=|\mathcal{T}_j|,
    \label{eq:agrisplat_gradient}
\end{equation}
where $\boldsymbol{\mu}_{j,t}^{2D}$ denotes the projected Gaussian center associated with offset $j$, and $\mathcal{T}_j$ is the set of iterations within the current statistics-accumulation interval in which its gradient is recorded.
CAR retains the original growth rules and their level-dependent thresholds, but considers offsets satisfying $0<g_j<\tau_{L(j)}$ and $n_j>n_{\min}$, where $\tau_{L(j)}$ is the original growth threshold at the parent anchor level $L(j)$.
The minimum observation requirement $n_{\min}$ excludes offsets with insufficient gradient observations, and only offsets at active octree levels are considered.

Each eligible offset proposes a 3D location that maps to an octree cell at the same level as its parent anchor:
\begin{equation}
    \mathbf{x}_j =
    \mathbf{a}_j+\mathbf{o}_j\odot\mathbf{s}_j,
    \qquad
    \mathbf{c}_j =
    \operatorname{round}
    \left(
        \frac{\mathbf{x}_j-\mathbf{p}_0}{q_{L(j)}}
    \right),
    \label{eq:agrisplat_cell}
\end{equation}
where $\mathbf{a}_j$ is the parent anchor position, $\mathbf{o}_j$ is its offset, $\mathbf{s}_j$ denotes the corresponding offset scaling, $\mathbf{p}_0$ is the octree origin, and $q_{L(j)}$ is the cell size at level $L(j)$.
The destination must be unoccupied and satisfy the original viewpoint-dependent admission conditions.
Occupancy is checked after standard anchor growth, including anchors introduced during the current round.
When multiple offsets propose the same cell at a given level, only the candidate with the highest gradient statistic is retained.
The remaining candidates are ranked by their original $g_j$ values without additional level normalization.

To distribute the available credit over successive growth rounds, CAR allocates a portion of the remaining credit at each round.
Each reinvestment window extends from a credited contribution-pruning round to the next contribution-pruning round or the end of anchor growth.
At each eligible growth round, the number of newly introduced anchors and the credit update are determined by
\begin{equation}
    m_t =
    \min\left(
        \left\lceil\frac{C_t}{K_t}\right\rceil,
        N_t
    \right),
    \qquad
    C_t \leftarrow C_t-m_t,
    \label{eq:agrisplat_reinvestment}
\end{equation}
where $C_t$ is the available credit, $K_t$ is the number of growth rounds remaining in the current window including the current round, and $N_t$ is the number of distinct eligible level--cell pairs after standard growth.
The $m_t$ highest-ranked level--cell pairs receive new anchors using the corresponding parent features and the standard same-level initialization of Octree-GS.
Only newly introduced anchors consume credit, while any unused balance is carried forward and supplemented by credit from subsequent contribution-pruning rounds.
Removals at the final pruning round do not generate credit, and any remaining credit is discarded when anchor growth ends.
CAR thus couples contribution-based pruning with budgeted growth at sub-threshold locations without imposing a fixed total anchor count.

\subsection{Asymmetric Directional Supervision}
\label{sec:agrisplat_ads}

The NVS benchmark reveals that existing reconstruction methods tend to under-reproduce gradient energy within crop regions, suggesting insufficient reproduction of local image contrast.
To address this observation, ADS introduces a one-sided image-space constraint that penalizes missing contrast along the reference image-gradient direction.
Rather than matching image gradients symmetrically, ADS focuses on the directional component of contrast that is insufficiently reproduced.
The supervision is applied after anchor growth and CAR reallocation are complete, allowing local contrast refinement with a fixed anchor topology rather than during ongoing anchor pruning and growth.

\myparagraph{Directional Contrast Deficit}
Let $I$ and $\hat{I}$ denote the reference and rendered RGB images, respectively.
To measure contrast in a single luminance channel, both images are converted using $Y=0.299R+0.587G+0.114B$, followed by horizontal and vertical forward differences.
At each image position $p$, the reference and rendered luminance gradients are denoted by $\mathbf{h}_p=\nabla Y(I)_p$ and $\mathbf{g}_p=\nabla Y(\hat{I})_p$, respectively.
The reference gradient direction and the local contrast deficit are defined as
\begin{equation}
    \mathbf{u}_p =
    \frac{\mathbf{h}_p}{\|\mathbf{h}_p\|_2},
    \qquad
    \ell_p =
    \max\left(
        0,\,
        \|\mathbf{h}_p\|_2 -
        \mathbf{g}_p^\top\mathbf{u}_p
    \right),
    \label{eq:agrisplat_deficit}
\end{equation}
where $\mathbf{u}_p=\mathbf{0}$ and $\ell_p=0$ when $\|\mathbf{h}_p\|_2=0$.
The ADS loss and training objective during the ADS stage are defined as
\begin{equation}
    \mathcal{L}_{\mathrm{ADS}}
    = \frac{1}{|\mathcal{D}|}
    \sum_{p\in\mathcal{D}}\ell_p,
    \quad
    \mathcal{L}
    = \mathcal{L}_{\mathrm{Octree\text{-}GS}}
    + \lambda_{\mathrm{ADS}}\mathcal{L}_{\mathrm{ADS}},
    \label{eq:agrisplat_ads}
\end{equation}
where $\mathcal{D}$ denotes the valid forward-difference positions over the full image, $\mathcal{L}_{\mathrm{Octree\text{-}GS}}$ is the photometric and regularization objective, and $\lambda_{\mathrm{ADS}}$ controls the ADS contribution.

The local deficit $\ell_p$ is the positive part of the difference between the reference gradient magnitude and the rendered gradient projected onto the reference direction.
This one-sided formulation penalizes insufficient reference-aligned contrast without introducing an additional penalty for excess contrast through the ADS term.
ADS is applied over the full training image without crop-region annotations.

Module-wise ablation studies and a comparison between symmetric and asymmetric gradient supervision are presented in Section~6.4, while details of hyperparameter selection are provided in the Supplementary Material S4.2.

%% file: sections/06_experiments.tex
\clearpage
\section{Experiments}
\subsection{Experimental Setup}
\label{sec:experimental_setup}

\begin{figure}[pos=h]
    \centering
    \includegraphics[width=\columnwidth]{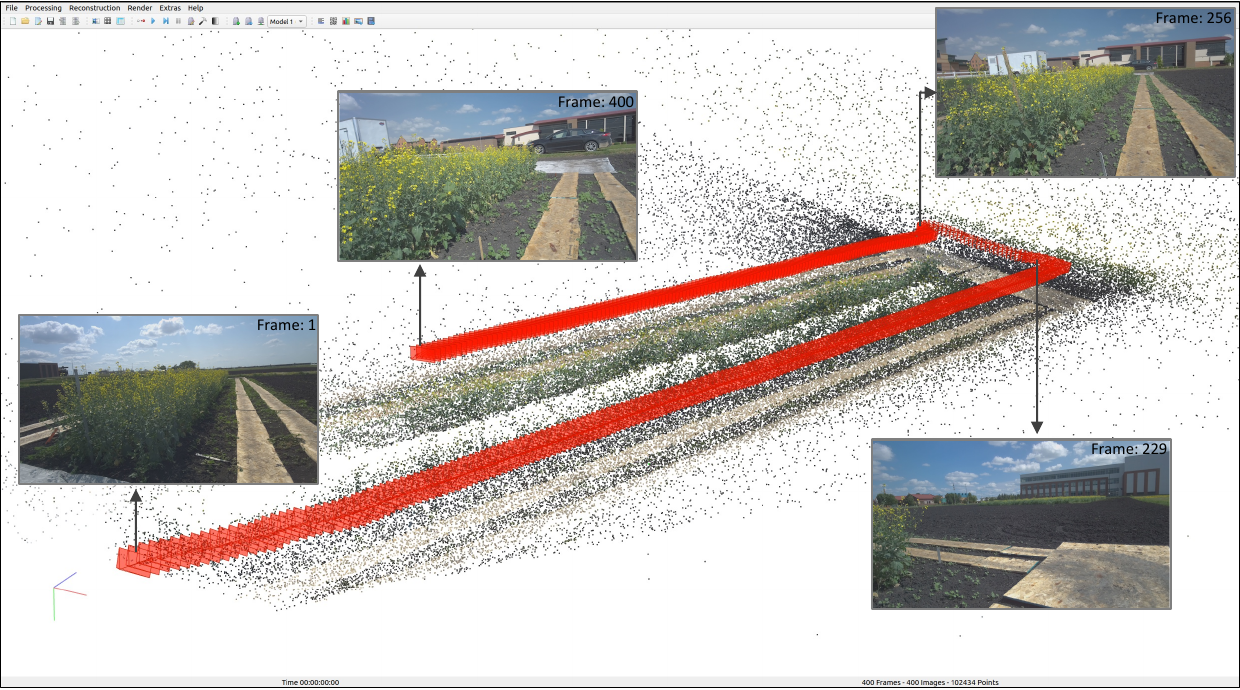}
    \caption{
        \textbf{Visualization of a representative scenario.}
        Spatial configuration of representative RGB observations, camera poses, and the initial point cloud from a Site 1 traversal.
    }
\vspace{-5pt}
\label{fig:benchmark_scene}    
\end{figure}

\myparagraph{Implementation Details}
AgriSplat builds upon Octree-GS and uses the same 40,000-iteration training schedule as the Octree-GS baseline in our experiments.
During initialization, visibility-based anchor filtering is relaxed to retain candidates selected by at least one training camera.
CAR performs Contribution Pruning and Gradient-Ranked Reinvestment alongside the anchor optimization process, with anchor growth and pruning ending at iteration 25,000 as in Octree-GS.
ADS is subsequently applied during the remaining 15,000 iterations, keeping the anchor topology fixed while continuing to optimize the trainable Gaussian parameters.
AgriSplat is evaluated using three random seeds (0, 1, and 2), and the reported results are averaged across these runs.
Additional hyperparameter settings and their analyses are provided in the Supplementary Material S4.2.

\begin{figure}[pos=t]
    \vspace{0pt}
    \centering
    \includegraphics[width=\columnwidth]{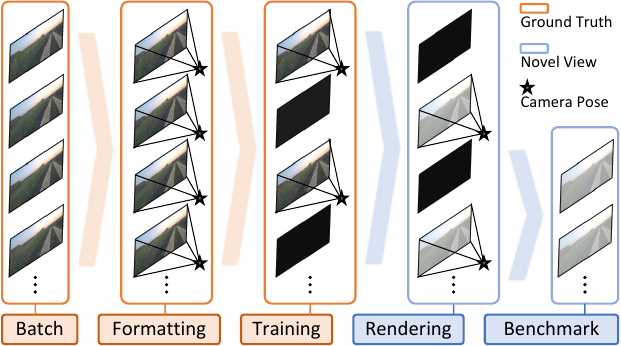}
    \caption{
        Benchmark workflow comprising frame preparation, COLMAP-compatible formatting, model training, novel-view rendering, and evaluation against reference images.
    }
\vspace{-10pt}
\label{fig:benchmark_workflow}
\end{figure}

\myparagraph{Benchmark Definition}
We establish a novel view synthesis (NVS) benchmark using AgriChrono to evaluate 3D reconstruction under varying natural illumination and crop growth conditions.
Seven scenarios are selected from the counterclockwise (CCW) traversals at Site 1: four \textit{Lighting Variance} sessions acquired at 06:00, 11:00, 16:00, and 21:00 on Day 19, and three \textit{Growth Span} sessions acquired at 06:00 on Days 6, 13, and 20.
For each scenario, we sample every seventh frame from the left RGB stream of ZED-L, obtaining 400 images at a resolution of $960 \times 540$.
Of these, 350 images are used for training, and 50 uniformly distributed images are held out for evaluation.
Each scenario is reconstructed independently using its training images and camera poses, with an initial point cloud generated by sampling 0.1\% of depth points from the training frames.
The benchmark thus evaluates held-out-view rendering within individual traversals rather than generalization to unseen acquisition dates or sites.
Figures~\ref{fig:benchmark_scene} and~\ref{fig:benchmark_workflow} illustrate a representative scenario and the overall benchmark workflow, respectively.

We compare AgriSplat with 15 baselines, comprising 11 Gaussian Splatting-based and four NeRF-based methods.
Baseline experiments use NerfBaselines v1.2.12~\citep{kulhanek2025nerfbaselines} with the default training configurations.

\myparagraph{Evaluation Protocol}
To evaluate the novel-view rendering quality of reconstructed agricultural scenes, we compare reference images and renderings at two spatial scopes: \textit{Whole Image} and \textit{Canola Region}.
For the Whole Image, we employ PSNR ($\mathrm{PSNR}_{W}$)~\citep{hore2010image}, SSIM ($\mathrm{SSIM}_{W}$)~\citep{wang2004image}, and LPIPS ($\mathrm{LPIPS}_{W}$)~\citep{zhang2018unreasonable}.
The Canola Region is evaluated using SSIM ($\mathrm{SSIM}_{C}$) and LPIPS ($\mathrm{LPIPS}_{C}$), based on masks manually annotated with SAM 3~\citep{carion2026sam3} across all 350 test images.
To further characterize fine-detail rendering within crop regions, we additionally employ the Gradient-Energy Ratio ($\mathrm{GER}_{C}$) and Edge F-score ($\mathrm{Edge\text{-}F}_{C}$).
The former compares gradient energy between reference and rendered images, with values closer to one indicating agreement, whereas the latter measures spatial edge correspondence within a 2-pixel tolerance, allowing minor misalignments while penalizing missing and spurious boundaries.

\begin{table}[pos=h]
    \centering
    \vspace{0pt}
    \caption{
        \textbf{Computational resource usage.}
        Mean training time and GPU memory usage across seven benchmark scenarios on an NVIDIA RTX 6000 Ada GPU.
        GPU memory represents the mean of per-scenario peak values.
        Baseline measurements were obtained using NerfBaselines~\citep{kulhanek2025nerfbaselines}.
    }
    \renewcommand{\arraystretch}{1.1}
    \setlength{\tabcolsep}{3pt}
    \small
    \adjustbox{max width=\columnwidth}{
        \begin{tabular}{l|r|r}
            \toprule
            \textbf{Method}
            & \textbf{Training Time}
            & \textbf{GPU Memory} \\
            \midrule

            \multicolumn{3}{l}{\textit{NeRF-based Methods}} \\
            \midrule
            SeaThru-NeRF~\citep{levy2023seathru} & 11\,h 51\,m 45\,s & 37,218\,MB \\
            Nerfacto~\citep{nerfstudio} & 6\,m 53\,s & 4,198\,MB \\
            Tetra-NeRF~\citep{kulhanek2023tetranerf} & 17\,h 53\,m 04\,s & 13,708\,MB \\
            Zip-NeRF~\citep{barron2023zipnerf} & 4\,h 27\,m 45\,s & 32,448\,MB \\
            \midrule

            \multicolumn{3}{l}{\textit{Gaussian Splatting-based Methods}} \\
            \midrule
            3DGS~\citep{kerbl20233d} & 6\,m 43\,s & 4,665\,MB \\
            Scaffold-GS~\citep{scaffoldgs} & 8\,m 25\,s & 4,785\,MB \\
            Taming 3DGS~\citep{Taming3DGS} & 4\,m 31\,s & 6,016\,MB \\
            3DGRT~\citep{3dgrt2024} & 27\,m 52\,s & 7,866\,MB \\
            H3DGS~\citep{hierarchicalgaussians24} & 38\,m 12\,s & 8,602\,MB \\
            GS-W~\citep{zhang2024GS-W} & 58\,m 32\,s & 7,027\,MB \\
            3D-MCMC~\citep{kheradmand20243d} & 4\,m 56\,s & 4,512\,MB \\
            PGSR~\citep{chen2024pgsr} & 7\,m 30\,s & 4,770\,MB \\
            gsplat~\citep{ye2025gsplat} & 5\,m 15\,s & 1,834\,MB \\
            3DGUT~\citep{wu20253dgut} & 7\,m 00\,s & 6,315\,MB \\
            Octree-GS~\citep{ren2024octree} & 13\,m 23\,s & 5,578\,MB \\
            \midrule
            \textbf{AgriSplat (Ours)} & 17\,m 58\,s & 6,258\,MB \\
            \bottomrule
        \end{tabular}
    }
    \label{tab:resource}
\end{table}

Whole-image metrics are computed over all 50 test frames per scenario, while Canola Region metrics are evaluated using valid annotated regions, with overall benchmark results obtained by equally weighting the seven scenario-level measurements. Table~\ref{tab:resource} reports the mean training time and mean per-scenario peak GPU memory usage for each method, measured on an NVIDIA RTX 6000 Ada GPU, while Supplementary Material S3 presents qualitative visualizations of the annotated Canola Regions, the formulations of GER and Edge F-score, and additional per-scenario quantitative results.

\begin{table*}[pos=t]
    \centering
    \caption{
        \textbf{Mean NVS performance over seven AgriChrono scenarios.}
        W and C denote Whole Image and Canola Region, respectively. Higher PSNR, SSIM, and Edge F-score and lower LPIPS are better; GER is evaluated by proximity to 1. Notably, strong Whole Image performance does not necessarily translate to equally strong Canola Region performance across the evaluated metrics. 
        \textbf{Bold} and \underline{underlined} values denote the best and second-best results, respectively.
    }
    \label{tab:nvs_overall}
    \renewcommand{\arraystretch}{1.1}
    \setlength{\tabcolsep}{6pt}
    \small
    \adjustbox{max width=\textwidth}{
        \begin{tabular}{lccccccc}
            \toprule
            & \multicolumn{3}{c}{Whole Image}
            & \multicolumn{4}{c}{Canola Region} \\
            \cmidrule(lr){2-4}\cmidrule(lr){5-8}
            Method & PSNR$_W\uparrow$ & SSIM$_W\uparrow$ & LPIPS$_W\downarrow$ & SSIM$_C\uparrow$ & LPIPS$_C\downarrow$ & GER$_C\rightarrow 1$ & EdgeF$_C\uparrow$ \\
            \midrule

            SeaThru-NeRF~\citep{levy2023seathru} & 23.49 & 0.527 & 0.671 & 0.336 & 0.850 & 0.024 & 0.021 \\
            Nerfacto~\citep{nerfstudio} & 22.06 & 0.529 & 0.543 & 0.337 & 0.767 & 0.047 & 0.100 \\
            Tetra-NeRF~\citep{kulhanek2023tetranerf} & 26.71 & 0.671 & 0.430 & 0.433 & 0.635 & 0.120 & 0.254 \\
            Zip-NeRF~\citep{barron2023zipnerf} & \textbf{28.26} & \underline{0.760} & \underline{0.246} & \textbf{0.560} & 0.375 & 0.378 & 0.637 \\
            \midrule
            3DGS~\citep{kerbl20233d} & 26.31 & 0.715 & 0.369 & 0.507 & 0.530 & 0.340 & 0.540 \\
            Scaffold-GS~\citep{scaffoldgs} & 24.61 & 0.684 & 0.367 & 0.467 & 0.530 & 0.394 & 0.586 \\
            Taming 3DGS~\citep{Taming3DGS} & 26.70 & 0.722 & 0.360 & 0.513 & 0.507 & 0.352 & 0.566 \\
            3DGRT~\citep{3dgrt2024} & 25.93 & 0.710 & 0.357 & 0.497 & 0.516 & 0.337 & 0.550 \\
            H3DGS~\citep{hierarchicalgaussians24} & 26.74 & 0.741 & 0.260 & 0.516 & \underline{0.363} & \underline{0.517} & 0.709 \\
            GS-W~\citep{zhang2024GS-W} & 21.03 & 0.582 & 0.575 & 0.375 & 0.791 & 0.087 & 0.173 \\
            3D-MCMC~\citep{kheradmand20243d} & 25.71 & 0.643 & 0.527 & 0.425 & 0.736 & 0.144 & 0.299 \\
            PGSR~\citep{chen2024pgsr} & 25.70 & 0.671 & 0.453 & 0.465 & 0.598 & 0.267 & 0.462 \\
            gsplat~\citep{ye2025gsplat} & 26.28 & 0.707 & 0.381 & 0.495 & 0.548 & 0.316 & 0.514 \\
            3DGUT~\citep{wu20253dgut} & 25.11 & 0.609 & 0.496 & 0.383 & 0.699 & 0.122 & 0.261 \\
            Octree-GS~\citep{ren2024octree} & 27.03 & 0.750 & 0.276 & 0.533 & 0.391 & 0.506 & \underline{0.712} \\
            \midrule
            \textbf{AgriSplat (Ours)} & \underline{27.24} & \textbf{0.779} & \textbf{0.187} & \textbf{0.560} & \textbf{0.273} & \textbf{0.847} & \textbf{0.797} \\
            \bottomrule
        \end{tabular}
    }
\end{table*}

\begin{table*}[pos=t]
    \centering
    \caption{
        \textbf{Scenario-wise Canola Region reconstruction performance.} Given the focus on crop reconstruction, we compare AgriSplat with four representative 3D reconstruction methods exhibiting competitive performance across crop-region metrics. The evaluation covers seven scenarios using SSIM, LPIPS, GER, and Edge F-score. Higher SSIM and Edge F-score and lower LPIPS indicate better performance, while GER is evaluated by proximity to 1. \textbf{Bold} and \underline{underlined} denote the best and second-best.
    }
    \label{tab:nvs_canola_comparison}
    \renewcommand{\arraystretch}{1.1}
    \setlength{\tabcolsep}{9pt}
    \small
    \adjustbox{max width=\textwidth}{
        \begin{tabular}{ll|rrrr|rrr}
            \toprule
            & & \multicolumn{4}{c|}{
                \textit{Lighting Variance (Day 19)}
            }
            & \multicolumn{3}{c}{
                \textit{Growth Span (06:00)}
            } \\
            \cmidrule(lr){3-6}
            \cmidrule(lr){7-9}
            \textbf{Method}
            & \textbf{Metric} & 06:00 & 11:00 & 16:00 & 21:00 & Day 6 & Day 13 & Day 20 \\
            \midrule

            \multirow{4}{*}{
                Zip-NeRF~\citep{barron2023zipnerf}
            }
            & SSIM$_C$ & 0.553 & \textbf{0.448} & \textbf{0.468} & 0.560 & \underline{0.620} & \underline{0.669} & \textbf{0.602} \\
            & LPIPS$_C$ & 0.482 & \underline{0.403} & 0.410 & 0.474 & \underline{0.238} & \underline{0.282} & \underline{0.336} \\
            & GER$_C$ & 0.253 & 0.367 & 0.362 & 0.244 & 0.555 & 0.454 & 0.412 \\
            & EdgeF$_C$ & 0.369 & 0.763 & 0.712 & 0.391 & 0.890 & 0.712 & 0.618 \\
            \midrule

            \multirow{4}{*}{
                Scaffold-GS~\citep{scaffoldgs}
            }
            & SSIM$_C$ & 0.512 & 0.318 & 0.392 & 0.532 & 0.516 & 0.554 & 0.447 \\
            & LPIPS$_C$ & 0.593 & 0.606 & 0.566 & 0.587 & 0.323 & 0.437 & 0.595 \\
            & GER$_C$ & 0.290 & 0.433 & 0.345 & 0.289 & 0.554 & 0.468 & 0.378 \\
            & EdgeF$_C$ & 0.328 & 0.771 & 0.676 & 0.317 & 0.870 & 0.648 & 0.492 \\
            \midrule

            \multirow{4}{*}{
                H3DGS~\citep{hierarchicalgaussians24}
            }
            & SSIM$_C$ & 0.574 & 0.366 & 0.414 & 0.574 & 0.547 & 0.615 & 0.523 \\
            & LPIPS$_C$ & \underline{0.377} & 0.435 & \underline{0.406} & \underline{0.393} & 0.253 & 0.300 & 0.376 \\
            & GER$_C$ & 0.440 & \underline{0.502} & \underline{0.511} & \underline{0.444} & 0.659 & \underline{0.562} & \underline{0.497} \\
            & EdgeF$_C$ & 0.525 & \underline{0.835} & 0.794 & \underline{0.521} & 0.901 & 0.736 & \underline{0.650} \\
            \midrule

            \multirow{4}{*}{
                Octree-GS~\citep{ren2024octree}
            }
            & SSIM$_C$ & \underline{0.589} & 0.390 & 0.429 & \underline{0.581} & 0.590 & 0.625 & 0.528 \\
            & LPIPS$_C$ & 0.393 & 0.469 & 0.435 & 0.428 & 0.268 & 0.325 & 0.419 \\
            & GER$_C$ & \underline{0.467} & 0.449 & 0.502 & 0.419 & \underline{0.674} & 0.558 & 0.472 \\
            & EdgeF$_C$ & \underline{0.564} & 0.819 & \underline{0.799} & 0.511 & \underline{0.905} & \underline{0.740} & 0.646 \\
            \midrule

            \multirow{4}{*}{
                \textbf{AgriSplat (Ours)}
            }
            & SSIM$_C$ & \textbf{0.609} & \underline{0.411} & \underline{0.443} & \textbf{0.596} & \textbf{0.634} & \textbf{0.670} & \underline{0.556} \\
            & LPIPS$_C$ & \textbf{0.277} & \textbf{0.325} & \textbf{0.327} & \textbf{0.291} & \textbf{0.186} & \textbf{0.213} & \textbf{0.288} \\
            & GER$_C$ & \textbf{0.785} & \textbf{0.824} & \textbf{0.798} & \textbf{0.793} & \textbf{0.977} & \textbf{0.916} & \textbf{0.838} \\
            & EdgeF$_C$ & \textbf{0.686} & \textbf{0.888} & \textbf{0.844} & \textbf{0.680} & \textbf{0.924} & \textbf{0.807} & \textbf{0.751} \\
            \bottomrule

        \end{tabular}
    }
\end{table*}

\subsection{Quantitative Results}
\label{subsec:quantitative_results}

The quantitative results and computational resource usage of the 15 baseline methods and AgriSplat are presented in Tables~\ref{tab:resource}--\ref{tab:nvs_canola_comparison}.
Table~\ref{tab:resource} reports training time and GPU memory usage, while Table~\ref{tab:nvs_overall} summarizes the average performance across seven scenarios using PSNR, SSIM, and LPIPS for the Whole Image, and SSIM, LPIPS, GER, and Edge F-score for the Canola Region.
Table~\ref{tab:nvs_canola_comparison} provides a scenario-wise comparison of AgriSplat with four representative reconstruction methods using the four Canola Region metrics.
Detailed scenario-wise results for all baselines are provided in the Supplementary Material S3.3.

\myparagraph{NeRF-based Baselines}
NeRF-based methods exhibit considerable differences in rendering quality and computational cost.
Nerfacto achieves relatively short training times but lower image similarity, whereas SeaThru-NeRF, designed to model image formation in scattering media, exhibits particularly low gradient energy and edge correspondence within the Canola Region.
Tetra-NeRF, which employs a tetrahedral scene representation, achieves comparatively higher image similarity but requires the longest training time among all baselines.
Zip-NeRF, which combines multi-scale representations with anti-aliasing, achieves the highest Whole Image PSNR and SSIM and the lowest LPIPS among the baseline methods, as well as the highest Canola Region SSIM among the baselines.
However, it requires substantially longer training times than most Gaussian Splatting methods, while achieving lower GER and Edge F-scores than Octree-GS, H3DGS, and AgriSplat, indicating differences between overall rendering fidelity and the reproduction of fine image details within crop regions.

\myparagraph{Gaussian-based Baselines}
Gaussian Splatting-based methods exhibit different characteristics in image similarity, computational efficiency, and the reconstruction of fine crop details across the seven benchmark scenarios.
Compared with the original 3DGS, Taming 3DGS achieves higher image similarity with shorter training times, while gsplat maintains comparable rendering quality with substantially lower GPU memory consumption.
In contrast, 3D-MCMC exhibits relatively lower image similarity under the evaluated default configuration.

In particular, methods employing structured and hierarchical Gaussian representations demonstrate strong performance within the Canola Region.
Scaffold-GS, which generates local Gaussians from anchor-based representations, Octree-GS, which employs multi-resolution anchors and level-of-detail selection, and H3DGS, which organizes Gaussians hierarchically, all rank among the top three baselines in GER and the top four in Edge F-score.
Among these methods, Octree-GS achieves high regional image similarity and the highest Edge F-score across all baselines, while H3DGS records the GER closest to one with a comparable Edge F-score.
Scaffold-GS also exhibits relatively higher performance in regional gradient energy and edge correspondence than in overall image similarity.
These trends motivate examining structured and hierarchical Gaussian representations for fine-detail reconstruction, although the effects of representation design cannot be isolated from other differences among the methods.

Meanwhile, PGSR, which incorporates planar geometric constraints, GS-W, which models appearance variations, and 3DGRT and 3DGUT, which employ different rendering strategies, exhibit distinct performance characteristics, including differences in training time.
In particular, 3DGRT outperforms 3DGUT in overall image similarity and regional edge correspondence, whereas GS-W shows relatively lower performance across both evaluation regions.

\myparagraph{Scenario-wise Analysis}
As shown in Table~\ref{tab:nvs_canola_comparison} and Supplementary Tables S6--S8, reconstruction performance varies across illumination and crop growth conditions.
For the Lighting Variance scenarios, visual inspection suggests increasing image brightness in the order of 21:00, 06:00, 16:00, and 11:00.
All five compared methods achieve lower Canola Region SSIM at 11:00 and 16:00 than at 06:00 and 21:00, suggesting sensitivity to acquisition-specific illumination conditions.
In contrast, their Edge F-scores consistently peak at 11:00.
This discrepancy may reflect differences in metric sensitivity, as SSIM evaluates luminance, contrast, and structural similarity, whereas Edge F-score measures spatial edge correspondence.
Stronger illumination may accentuate leaf boundaries and shadow-induced intensity transitions, although other acquisition conditions may also contribute to the observed differences.

Across the Growth Span scenarios, all five methods exhibit increasing Canola Region LPIPS and decreasing GER and Edge F-scores from Day 6 to Day 20, while SSIM peaks on Day 13 before declining below its Day 6 value.
The concurrent decline in GER and Edge F-score suggests weaker contrast reproduction and less accurate boundary correspondence at later stages of crop growth.
These trends are consistent with increasing reconstruction difficulty at later growth stages, potentially associated with greater structural complexity, leaf overlap, and occlusion, although other date-specific conditions may also contribute.
Furthermore, GER remains below one across all baselines and scenarios, indicating insufficient reproduction of crop-region gradient energy.
These findings highlight agricultural reconstruction challenges under varying illumination and crop growth conditions and the value of region-specific evaluation.

\clearpage
\begin{figure*}[pos=t]
    \centering
    \includegraphics[width=\textwidth]{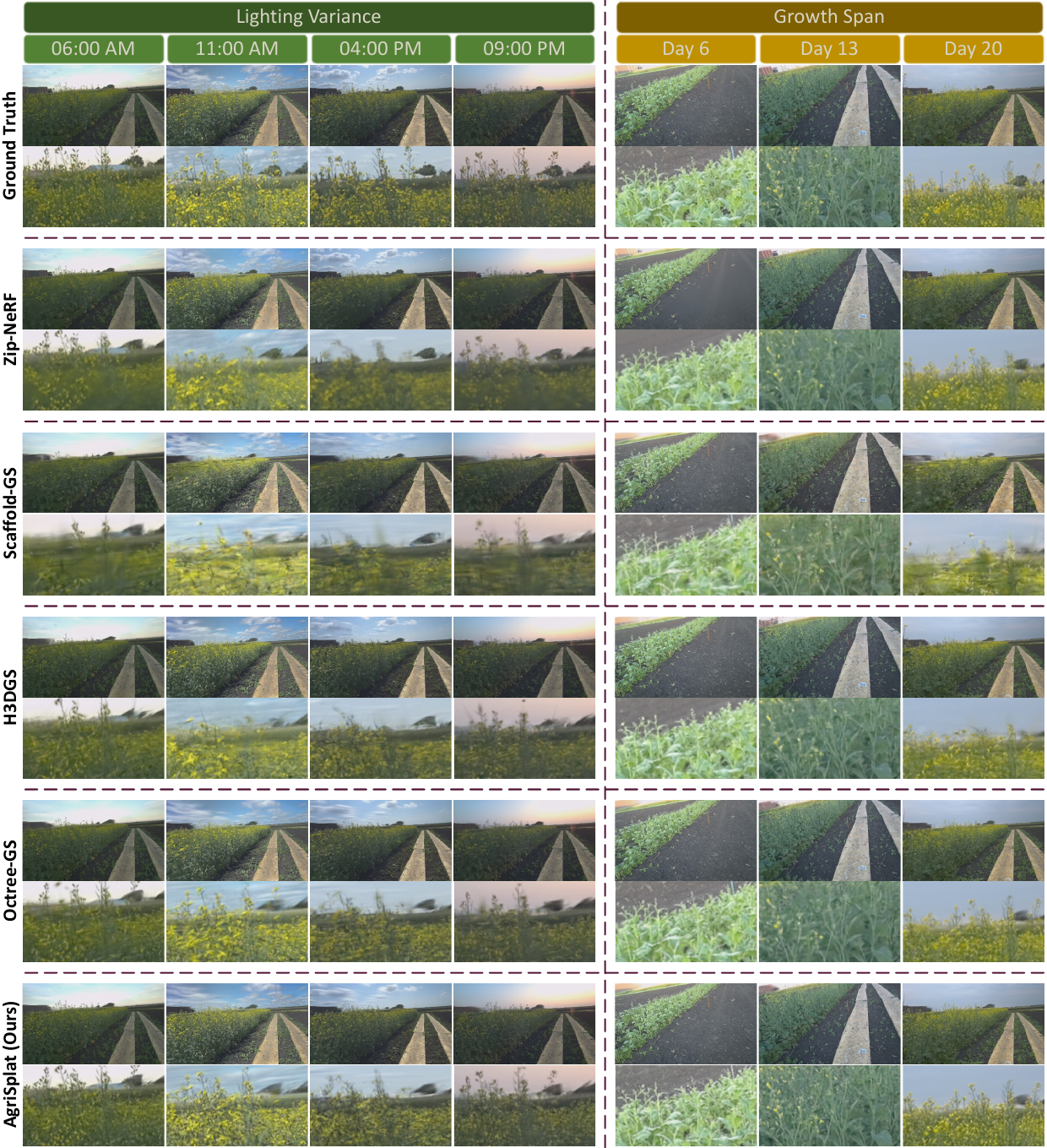}
    \caption{
        \textbf{Qualitative comparison of novel-view synthesis results across seven AgriChrono scenarios.}
        The \textit{Lighting Variance} group (left) presents four recording sessions acquired at 06:00, 11:00, 16:00, and 21:00 on Day 19, illustrating variations in natural illumination.
        The \textit{Growth Span} group (right) presents three sessions acquired at 06:00 on Days 6, 13, and 20, illustrating changes in crop appearance and structure over time.
        The top block shows the ground-truth reference images, followed by renderings from Zip-NeRF~\citep{barron2023zipnerf}, Scaffold-GS~\citep{scaffoldgs}, H3DGS~\citep{hierarchicalgaussians24}, Octree-GS~\citep{ren2024octree}, and the proposed AgriSplat.
        For each method, the upper row displays the Whole Image, while the lower row provides a corresponding zoomed-in view of the Canola Region.
        While the Whole Image views illustrate overall scene reconstruction and illumination reproduction, the enlarged views highlight differences in fine-detail rendering, including thin stems, overlapping leaves, flowering structures, and local boundary clarity.
        AgriSplat generally produces more distinct vegetation details than the compared baselines across the evaluated conditions.
    }
    \label{fig:qualitative_comparison}
\end{figure*}

\clearpage
\myparagraph{AgriSplat Analysis}
As shown in Table~\ref{tab:nvs_overall}, AgriSplat achieves the best average performance in six of the seven evaluation metrics, including Whole Image SSIM and LPIPS, and Canola Region SSIM, LPIPS, GER, and Edge F-score.
It also ranks second in Whole Image PSNR and achieves a Canola Region SSIM comparable to Zip-NeRF.
Compared with Octree-GS, AgriSplat improves all evaluated metrics, with particularly notable gains in regional gradient energy and edge correspondence.
The concurrent improvements in regional LPIPS and Edge F-score indicate that higher gradient energy is accompanied by better perceptual similarity and spatial edge correspondence, rather than an increase in gradient energy alone.

The scenario-wise results in Table~\ref{tab:nvs_canola_comparison} further show that AgriSplat achieves the lowest Canola Region LPIPS, the GER closest to one, and the highest Edge F-score in all seven scenarios among the five compared methods.
For regional SSIM, AgriSplat ranks first in four scenarios and second in the remaining three.
These results indicate that AgriSplat consistently improves fine-detail reproduction across the evaluated illumination and crop-growth conditions, while also improving overall perceptual rendering quality.
The individual contributions of CAR and ADS are further examined through the ablation studies in Section~\ref{subsec:ablation_study}.

\subsection{Qualitative Results}
\label{subsec:qualitative_results}

Figure~\ref{fig:qualitative_comparison} compares AgriSplat with four representative baselines across the seven evaluation scenarios.
For each method, the upper row presents whole-image renderings at $960 \times 540$ resolution, while the lower row shows corresponding enlarged Canola Regions.
At the whole-image scale, the compared methods generally reproduce the crop-row layout and overall scene appearance under different acquisition conditions.
However, the enlarged views reveal more pronounced differences in fine-detail reproduction.
Several baselines exhibit smoothing and ambiguous boundaries around thin stems, overlapping leaves, and flowering structures, whereas AgriSplat generally preserves more distinct local boundaries and better separates adjacent vegetation elements.
This contrast between plausible whole-image renderings and less faithful crop-region details is consistent with the differences observed between Whole Image and Canola Region metrics, highlighting the importance of region-specific evaluation.

The Growth Span examples show a progressive decline in crop-region rendering quality from Day 6 to Day 20, with increasingly indistinct vegetation details as crop structures become more complex.
This visual trend agrees with the scenario-wise quantitative results, where crop-region LPIPS increases, GER moves farther from one, and Edge F-score decreases across the compared methods.
In contrast, the Lighting Variance examples show no clear monotonic relationship between apparent image brightness and fine-detail rendering quality.
Instead, the visibility and reproduction of local structures vary with illumination, shadows, and scene content, consistent with the metric-dependent variations observed in the quantitative analysis.

The more distinct crop boundaries and vegetation details in AgriSplat renderings are consistent with its improvements in Canola Region LPIPS, GER, and Edge F-score.
These qualitative comparisons provide visual evidence of improved fine-detail reproduction across the evaluated conditions, supporting the findings from multiple region-specific metrics rather than gradient-energy agreement alone.

\subsection{Ablation Study}
\label{subsec:ablation_study}

We conduct two ablation studies to examine the effects of the proposed components and asymmetric directional supervision.
The module-wise ablation evaluates Contribution Pruning (CP), Gradient-Ranked Reinvestment (GR), and Asymmetric Directional Supervision (ADS) under a shared visibility-relaxed initialization setting (NF), while the second study compares symmetric and asymmetric versions of the aligned gradient penalty.
Unlike the main quantitative results averaged over three random seeds, all ablation configurations are evaluated using a single seed (seed 0) across the seven AgriChrono scenarios.

\begin{table}[pos=h]
    \centering
    \caption{
        \textbf{Module-wise ablation of AgriSplat.}
        All configurations use visibility-relaxed initialization (NF).
        CP, GR, and ADS denote Contribution Pruning,
        Gradient-Ranked Reinvestment, and Asymmetric Directional
        Supervision, respectively.
        Full denotes the proposed AgriSplat, combining NF, CP, GR, and ADS.
        Results are averaged over seven scenarios using seed 0.
        $N$ denotes the mean anchor count in thousands.
        \textbf{Bold} and \underline{underlined} indicate the
        best and second-best results, respectively.
    }
    \label{tab:ablation_modules}
    \renewcommand{\arraystretch}{1.1}
    \setlength{\tabcolsep}{5pt}
    \small
    \adjustbox{max width=\columnwidth}{
        \begin{tabular}{lccccc}
            \toprule
            Metric & NF & NF+CP & NF+CP+GR & NF+ADS & Full \\
            \midrule
            PSNR$_W$ & \underline{27.41} & 27.37 & \textbf{27.55} & 27.10 & 27.24 \\
            SSIM$_W$ & 0.778 & 0.776 & \textbf{0.784} & 0.772 & \underline{0.779} \\
            LPIPS$_W$ & 0.210 & 0.219 & \underline{0.198} & 0.202 & \textbf{0.190} \\
            \midrule
            SSIM$_C$ & 0.555 & 0.555 & \textbf{0.568} & 0.545 & \underline{0.559} \\
            LPIPS$_C$ & 0.317 & 0.326 & \underline{0.289} & 0.304 & \textbf{0.277} \\
            GER$_C$ & 0.600 & 0.587 & 0.642 & \underline{0.780} & \textbf{0.837} \\
            EdgeF$_C$ & 0.759 & 0.754 & 0.778 & \underline{0.782} & \textbf{0.796} \\
            \midrule
            $N$ (k) & 785.8 & \textbf{455.6} & \underline{631.0} & 785.8 & \underline{631.0} \\
            \bottomrule
        \end{tabular}
    }
\end{table}

\myparagraph{Module-wise Ablation}
Table~\ref{tab:ablation_modules} summarizes the effects of individual components on rendering quality and anchor count.
Introducing CP reduces the anchor count by approximately 42\,\% with comparatively small changes in rendering quality, while adding GR improves image similarity by reinvesting the available credit into eligible sub-threshold candidates.
The resulting CP+GR configuration achieves the highest Whole Image PSNR and SSIM and Canola Region SSIM among the evaluated ablation configurations, using approximately 20\,\% fewer anchors than NF alone.
Adding ADS improves LPIPS, GER, and Edge F-score, although some reductions in PSNR and SSIM are observed.
The full AgriSplat configuration achieves the best LPIPS, GER, and Edge F-score among the evaluated variants.
As ADS operates after anchor growth and reallocation, these improvements are obtained without increasing the anchor count.
These findings suggest complementary roles for CAR and ADS, with CAR reallocating representation capacity and ADS subsequently refining local contrast within the resulting fixed anchor topology.

\begin{table}[pos=h]
    \centering
    \caption{
        \textbf{Symmetric versus asymmetric gradient supervision.}
        Both variants use the same CAR configuration and training
        settings, differing only in the symmetric absolute penalty
        versus the proposed one-sided hinge penalty.
        Results are averaged over seven scenarios using seed 0.
        \textbf{Bold} and \underline{underlined} indicate the
        best and second-best results, respectively.
    }
    \label{tab:ablation_ads}
    \renewcommand{\arraystretch}{1.1}
    \setlength{\tabcolsep}{17pt}
    \small
    \adjustbox{max width=\columnwidth}{
        \begin{tabular}{lcc}
            \toprule
            Metric & Symmetric & Asymmetric (Ours) \\
            \midrule
            PSNR$_W$ & \textbf{27.42} & \underline{27.24} \\
            SSIM$_W$ & \textbf{0.782} & \underline{0.779} \\
            LPIPS$_W$ & \underline{0.195} & \textbf{0.190} \\
            \midrule
            SSIM$_C$ & \textbf{0.565} & \underline{0.559} \\
            LPIPS$_C$ & \underline{0.286} & \textbf{0.277} \\
            GER$_C$ & \underline{0.708} & \textbf{0.837} \\
            EdgeF$_C$ & \underline{0.788} & \textbf{0.796} \\
            \bottomrule
        \end{tabular}
    }
\end{table}

\myparagraph{Symmetric versus Asymmetric Supervision}
Table~\ref{tab:ablation_ads} compares the one-sided ADS formulation with a symmetric version of the same aligned gradient penalty.
The symmetric variant replaces the one-sided hinge penalty with the absolute value of the aligned difference,
$\ell_p^{\mathrm{sym}}=\left|\|\mathbf{h}_p\|_2-\mathbf{g}_p^\top\mathbf{u}_p\right|$,
thereby penalizing both insufficient and excess reference-aligned contrast.
All other settings, including the supervision stage, loss weight, and initialization checkpoint, are kept unchanged.
Under the evaluated settings, the asymmetric formulation achieves lower Whole Image and Canola Region LPIPS and better Canola Region GER and Edge F-score, at the cost of slightly lower Whole Image PSNR and SSIM and Canola Region SSIM.
This trade-off is consistent with the ADS objective, which penalizes insufficient reference-aligned contrast without introducing an additional penalty for excess contrast through the ADS term.

Additional qualitative ablation comparisons using three cropped regions from the Day 20 scenario are provided in the Supplementary Material S4.1.

%% file: sections/07_conclusion.tex
\section{Conclusion}
\label{sec:conclusion}

This study presented AgriChrono, a longitudinal dataset and benchmark for novel-view synthesis of field-grown crops, supported by the AgriChrobot robotic acquisition platform.
The primary Site 1 collection comprises 160 canola recording sessions acquired over approximately one month, and the benchmark establishes seven evaluation scenarios spanning different natural illumination conditions and crop-development stages.
Evaluation of 15 NeRF- and Gaussian Splatting-based methods showed that competitive whole-image rendering quality does not necessarily translate into faithful reproduction of fine crop details, highlighting the importance of region-specific evaluation.
Motivated by these findings, we introduced AgriSplat, which combines contribution-guided anchor reallocation with subsequent asymmetric directional supervision.
Compared with Octree-GS, AgriSplat improved whole-image quality and crop-region perceptual similarity, gradient-energy agreement, and edge correspondence. Ablations supported the complementary roles of capacity reallocation and contrast refinement, with the latter performed under a fixed anchor topology.

The benchmark focuses on held-out-view rendering within individual acquisition sessions, and the results demonstrate consistent improvements in fine-detail reconstruction across the evaluated illumination conditions and crop-development stages.
Improved reproduction of fine crop structures, such as thin stems, overlapping leaves, and flowers, may enable more detailed observation of plant morphology for future image-based phenotyping applications.
Future work may extend evaluation to additional crop types and field environments, investigate reconstruction across acquisition dates, and assess the utility of AgriSplat reconstructions for crop phenotyping.
Together, AgriChrono and AgriSplat provide a data resource, an evaluation framework, and a reconstruction method for advancing fine-detail reconstruction of agricultural scenes under field conditions.

%% file: supplementary/main_supplementary.tex
% ============================================
% Supplementary Material
% ============================================
\clearpage
\onecolumn

% Page layout
\newgeometry{margin=22mm}

% Synchronize Elsevier caption widths
\setlength{\FullWidth}{\textwidth}

% Font and paragraph formatting
\fontsize{11pt}{13.6pt}\selectfont
\setlength{\parskip}{3pt}
\setlength{\parindent}{1em}

% Reset counters
\setcounter{section}{0}
\setcounter{subsection}{0}
\setcounter{figure}{0}
\setcounter{table}{0}
\setcounter{equation}{0}

% Supplementary numbering
\renewcommand{\thesection}{S\arabic{section}}
\renewcommand{\thesubsection}{S\arabic{section}.\arabic{subsection}}
\renewcommand{\thefigure}{S\arabic{figure}}
\renewcommand{\thetable}{S\arabic{table}}
\renewcommand{\theequation}{S\arabic{equation}}

% Unique hyperref anchors
\renewcommand{\theHsection}{supp.\arabic{section}}
\renewcommand{\theHsubsection}{supp.\arabic{section}.\arabic{subsection}}
\renewcommand{\theHfigure}{supp.\arabic{figure}}
\renewcommand{\theHtable}{supp.\arabic{table}}
\renewcommand{\theHequation}{supp.\arabic{equation}}

% Paragraph heading (define only if absent)
\providecommand{\myparagraph}[1]{%
    \par\addvspace{3pt}%
    \noindent\textbf{#1.}\enspace\ignorespaces
}

% Start supplementary-only table of contents
\startcontents[supp]

% Supplementary title
\begin{center}
    {\LARGE\bfseries Supplementary Material for\par}
    \vspace{3pt}
    {\LARGE
    AgriChrono: A Longitudinal Dataset and Benchmark \\
    for Novel-View Synthesis of Field-Grown Crops\par}
    \vspace{12pt}
    Jaehwan Jeong et al.
\end{center}

\vspace{16pt}

% Supplementary-only contents
\begin{center}
    {\large\bfseries Contents}
\end{center}
\printcontents[supp]{}{1}{}{}

\clearpage

% Supplementary sections
\input{supplementary/sections/S01_agrichrobot}
\input{supplementary/sections/S02_agrichrono}
\input{supplementary/sections/S03_nvs_benchmark}
\input{supplementary/sections/S04_acquisition_quality}

\stopcontents[supp]

%% file: supplementary/sections/S01_agrichrobot.tex
\section{AgriChrobot: Data Collection Platform}
\label{supp:agrichrobot}

\subsection{Hardware Configuration and Mechanical Design}
\label{supp:robot_hardware}

\begin{table*}[pos=h]
    \centering
    \caption{
        Principal hardware components and their functions in field data acquisition.
    }
    \label{tab:supp_robot_hardware}
    \small
    \renewcommand{\arraystretch}{1.15}
    \adjustbox{max width=\textwidth}{
        \begin{tabular}{@{}p{0.35\textwidth}p{0.57\textwidth}@{}}
            \toprule
            \textbf{Subsystem} & \textbf{Configuration and function} \\
            \midrule
            Mobile base & AgileX Scout 2.0 UGV for field traversal \\
            Stereo cameras & Two ZED X units with 4-mm lenses \\
            Image acquisition & 1920 $\times$ 1080 pixels per RGB view; approximately 15 frames/s \\
            Onboard computer & NVIDIA Jetson AGX Orin (64~GB) \\
            Camera mounting & Adjustable three-axis mounts with passive vibration isolation \\
            Communication & 5G router and omnidirectional antenna \\
            Operator camera & OBSBOT PTZ camera for remote observation \\
            Auxiliary sensor & Livox Mid-360 LiDAR \\
            Power and storage & DC--DC conversion modules and local data storage \\
            Mechanical assembly & Custom sensor frame, protective chassis, and sun shield \\
            \bottomrule
        \end{tabular}
    }
\end{table*}

The custom sensor frame and protective chassis were designed, fabricated, and assembled for installation on the AgileX Scout 2.0 UGV. The three-tier assembly accommodates the sensing, computing, power conversion, and communication components summarized in Table~\ref{tab:supp_robot_hardware}.

The two ZED X stereo cameras are mounted on independently adjustable three-axis supports incorporating passive vibration isolation. Their forward-oblique orientations provide different viewing directions during crop-row traversal. The mechanical assembly includes protective aluminum--acrylic panels and a sun shield for outdoor deployment. Power is distributed through DC--DC conversion modules supplying the required voltages to the onboard computing and sensing devices, while local storage is used for camera recordings and associated sensor data. Mechanical drawings, assembly information, and additional component specifications are provided in the AgriChrono repository.

\subsection{Camera Calibration}
\label{supp:camera_calibration}

\subsubsection{Within-Unit Stereo Calibration}
\label{supp:stereo_calibration}

The acquisition system uses two ZED X stereo units, denoted ZED-L and ZED-R. Each contains left and right image sensors and uses its calibration parameters for image rectification and depth estimation. The left image sensor of each stereo unit serves as its reference camera. At a resolution of $1920 \times 1080$, the intrinsic matrices of the left reference cameras of ZED-L and ZED-R are

\vspace{-2pt}
\begin{equation}
\label{eq:supp_stereo_intrinsics}
\mathbf{K}_{L} =
\begin{bmatrix}
1258.97 & 0 & 916.48 \\
0 & 1258.97 & 553.83 \\
0 & 0 & 1
\end{bmatrix},
\qquad
\mathbf{K}_{R} =
\begin{bmatrix}
1261.83 & 0 & 978.86 \\
0 & 1261.83 & 535.06 \\
0 & 0 & 1
\end{bmatrix}.
\end{equation}

Here, $f_x$ and $f_y$ denote the focal lengths in pixels, and $c_x$ and $c_y$ denote the principal-point coordinates. The corresponding right-to-left transformations within the two stereo units are

\vspace{-2pt}
\begin{equation}
\label{eq:supp_stereo_extrinsics}
\mathbf{T}^{L}_{\ell\leftarrow r} =
\begin{bmatrix}
1 & 0 & 0 & 0.120009026 \\
0 & 1 & 0 & 0 \\
0 & 0 & 1 & 0 \\
0 & 0 & 0 & 1
\end{bmatrix},
\qquad
\mathbf{T}^{R}_{\ell\leftarrow r} =
\begin{bmatrix}
1 & 0 & 0 & 0.120212349 \\
0 & 1 & 0 & 0 \\
0 & 0 & 1 & 0 \\
0 & 0 & 0 & 1
\end{bmatrix}.
\end{equation}

The translation components are expressed in metres, corresponding to stereo baselines of approximately $120.009$~mm and $120.212$~mm for ZED-L and ZED-R, respectively. These transformations describe the relative geometry between the left and right image sensors within each stereo unit.

\subsubsection{Inter-Unit Extrinsic Calibration}
\label{supp:intercamera_extrinsics}

The relative extrinsic transformation between the two ZED X stereo units was estimated using the multi-sensor calibration method of \citet{zuniga2019multisensor}. This method combines sensor motion with ground-plane constraints to estimate the six-degree-of-freedom transformation between sensors.

For the inter-unit calibration, the left image sensor of each stereo unit is defined as its reference camera. Let $L$ and $R$ denote the reference cameras of ZED-L and ZED-R, respectively. The transformation $\mathbf{T}_{L\leftarrow R}$ maps points from the ZED-R reference frame into the ZED-L reference frame. The estimated transformation obtained from the second recording group is

\begin{equation}
\label{eq:supp_intercamera_extrinsic}
\begin{bmatrix}
\mathbf{p}_{L}\\
1
\end{bmatrix}
=
\mathbf{T}_{L\leftarrow R}
\begin{bmatrix}
\mathbf{p}_{R}\\
1
\end{bmatrix},
\qquad
\widehat{\mathbf{T}}_{L\leftarrow R}
=
\begin{bmatrix}
 0.629380 & -0.007114 &  0.777065 &  0.179913 \\
 0.006195 &  0.999972 &  0.004137 & -0.014167 \\
-0.777073 &  0.002210 &  0.629407 & -0.045551 \\
 0        &  0        &  0        &  1
\end{bmatrix}.
\end{equation}

Translation components are expressed in metres. The relative rotation angle is approximately $50.99^\circ$, with a translation component of approximately $0.180$~m along the first axis of the ZED-L reference frame. The transformation describes the relative geometry between the two stereo units for the recording group.

\subsection{Embedded Software and Data Acquisition}
\label{supp:robot_software}

\begin{table*}[pos=h]
    \centering
    \caption{
        Principal software modules used for AgriChrobot operation and data acquisition.
    }
    \label{tab:supp_software_modules}
    \small
    \renewcommand{\arraystretch}{1.15}
    \adjustbox{max width=\textwidth}{
        \begin{tabular}{@{}p{0.35\textwidth}p{0.57\textwidth}@{}}
            \toprule
            \textbf{Module} & \textbf{Function} \\
            \midrule
            \texttt{1\_janus-streaming} & Remote interface, video streaming, and recording control \\
            \texttt{2\_scout-ros2-control} & ROS~2 vehicle control and CAN communication \\
            \texttt{3\_zed-data-tools} & Stereo-camera data acquisition and recording \\
            \texttt{4\_livox-data-tools} & Auxiliary LiDAR data recording \\
            \texttt{5\_systemd} & Local storage management and backup services \\
            \texttt{6\_network} & Network initialization and remote connectivity \\
            \bottomrule
        \end{tabular}
    }
\end{table*}

\subsubsection{Software Environment and System Modules}
\label{supp:software_environment}

The software was deployed on an NVIDIA Jetson AGX Orin (64~GB) running JetPack~6.2. The environment includes ROS~2 Humble for robot control and ZED SDK~5.0.2 for stereo-camera acquisition. Vehicle control is implemented through the ROS~2 interface and CAN-bus communication.

The remote operating interface uses Flask and Socket.IO, while video transmission is implemented using GStreamer and Janus WebRTC Gateway. The software is organized into modules for remote operation, vehicle control, stereo-camera recording, auxiliary sensor logging, data management, and network connectivity. Table~\ref{tab:supp_software_modules} summarizes the principal software modules and their functions.

\subsubsection{Remote Operation and Recording Control}
\label{supp:remote_operation}

The browser-based operating interface integrates real-time video feedback, robot motion control, PTZ camera control, system monitoring, and recording management. Operator commands are processed by the Flask and Socket.IO backend and transmitted to the corresponding control processes. Vehicle movement commands are handled by ROS~2 through the Scout control interface.

\clearpage
For remote video feedback, the GStreamer pipeline uses hardware-accelerated encoding on the Jetson platform. The deployed operator stream uses approximately $640 \times 360$ resolution at 15 frames/s with a nominal bitrate of 300~kb/s. This stream is separate from the full-resolution stereo-camera recordings used for the dataset. The interface also includes a zero-velocity command mechanism triggered by browser disconnection or loss of control focus.

\subsubsection{Data Recording and Processing}
\label{supp:data_recording}

The data acquisition workflow separates field recording from offline data preparation. During field acquisition, the two ZED X units record stereo-camera streams in \texttt{.svo2} containers at approximately 15 frames/s. The sensor metadata and timestamps are stored for subsequent processing. A shared session-start time reference is generated when recording is initiated, allowing the recording processes to organize data from the same acquisition session.

After field collection, the stored camera streams are processed to export four RGB image streams, two stereo-derived depth maps, and corresponding frame-indexed metadata, including VIO-estimated camera poses. The processed observations are organized into session-based directories for the AgriChrono dataset.

\subsubsection{Data Management}
\label{supp:data_management}

The onboard software includes services for local storage management, data backup, and remote network access. Recorded data can be transferred to external storage and the project data repository after field acquisition. The software implementation, deployment dependencies, and operating procedures are available in the public AgriChrono project repository.

%% file: supplementary/sections/S02_agrichrono.tex
\section{AgriChrono: Dataset Documentation}
\label{supp:agrichrono}

\subsection{Collection Schedule}
\label{supp:dataset_schedule}

\newcommand{\agcalday}[4]{%
\begin{tabular}[t]{@{}c@{}}
\textbf{#1}\\[-1pt]
\strut #2\\[-1pt]
\strut #3\\[-1pt]
\strut #4
\end{tabular}%
}

\setlength{\columnsep}{20pt}
\begin{wraptable}{r}{0.47\columnwidth}
\centering
\vspace{-60pt}
\caption{\textbf{AgriChrono collection calendar.}}
\vspace{3pt}
\label{tab:supp_collection_calendar}
\scriptsize
\setlength{\tabcolsep}{1.5pt}
\renewcommand{\arraystretch}{0.9}
\resizebox{\linewidth}{!}{%
\begin{tabular}{@{}ccccccc@{}}
\toprule
    \textbf{Sun} & \textbf{Mon} & \textbf{Tue} & \textbf{Wed} & \textbf{Thu} & \textbf{Fri} & \textbf{Sat} \\
    \midrule
    \agcalday{}{}{}{} & \agcalday{}{}{}{} & \agcalday{Jul 1}{}{}{} & \agcalday{2}{S1 (4)}{S2 (1)}{S3 (1)} & \agcalday{3}{S1 (3)}{}{S3 (1)} & \agcalday{4}{S1 (3)}{}{} & \agcalday{5}{S1 (2)}{}{} \\
    \midrule
    \agcalday{6}{S1 (4)}{}{} & \agcalday{7}{S1 (4)}{}{} & \agcalday{8}{S1 (4)}{}{} & \agcalday{9}{S1 (4)}{}{} & \agcalday{10}{S1 (3)}{S2 (1)}{S3 (1)} & \agcalday{11}{S1 (4)}{}{} & \agcalday{12}{S1 (4)}{}{} \\
    \midrule
    \agcalday{13}{S1 (4)}{S2 (1)}{S3 (1)} & \agcalday{14}{S1 (4)}{}{} & \agcalday{15}{S1 (4)}{}{} & \agcalday{16}{S1 (4)}{}{} & \agcalday{17}{S1 (4)}{S2 (1)}{S3 (1)} & \agcalday{18}{S1 (4)}{}{} & \agcalday{19}{S1 (4)}{}{} \\
    \midrule
    \agcalday{20}{S1 (4)}{}{} & \agcalday{21}{S1 (2)}{S2 (1)}{S3 (1)} & \agcalday{22}{S1 (1)}{}{} & \agcalday{23}{}{}{} & \agcalday{24}{S1 (1)}{}{} & \agcalday{25}{}{}{} & \agcalday{26}{}{}{} \\
    \midrule
    \agcalday{27}{}{}{} & \agcalday{28}{S1 (1)}{S2 (1)}{S3 (1)} & \agcalday{29}{S1 (1)}{}{} & \agcalday{30}{S1 (1)}{}{} & \agcalday{31}{S1 (1)}{}{} & \agcalday{Aug 1}{S1 (1)}{S2 (1)}{S3 (1)} & \agcalday{}{}{}{} \\
    \bottomrule
    \end{tabular}%
    }
\vspace{-50pt}
\end{wraptable}

Table~\ref{tab:supp_collection_calendar} presents the acquisition calendar from July~2 to August~1, 2025. During Phase~1 (July~2--21), data collection at Site~1 was scheduled four times daily. Phase~2 (July~22--August~1) involved less frequent acquisitions as crop growth slowed. The actual number of recordings varied with field conditions, including rainfall and muddy ground.

Each Site~1 collection round consisted of one clockwise (CW) and one counterclockwise (CCW) traversal. The calendar records 73 collection rounds during Phase~1 and seven additional rounds during Phase~2, yielding 80 rounds and 160 directional recording sessions. The calendar also documents seven sessions at Site~2 and eight sessions at Site~3.

%%%%%%%%%%%%%%%%%%%%%%%%%%%%%%%%%%%%%%%%%%%%
\subsection{Additional Acquisition Sites}
\label{supp:dataset_additional_sites}

In addition to Site~1, AgriChrobot was deployed at two agricultural trial sites, as shown in Figure~3 of the main paper.

\clearpage

\myparagraph{Site 2: Canola Genotype Trial}
Site~2 was a canola genotype trial containing 44 genotypes distributed across 11 blocks. The plots measured approximately $13.4 \times 0.91$~m, with planting conducted on May~30, 2025. The field contained canola plants representing different genotypes and variations in plant appearance and structure. Seven recording sessions were acquired at this site.

\myparagraph{Site 3: Flax Trial}
Site~3 was a flax (\textit{Gold ND}) trial comprising 16 plots arranged in a $4 \times 4$ layout. Each plot measured $2.44 \times 0.91$~m, with planting conducted on May~30, 2025. The field included herbicide-treated and manually weeded blocks, providing observations of flax and surrounding vegetation under different cultivation treatments. Eight sessions were acquired.

Table~\ref{tab:supp_full_dataset_stats} summarizes the AgriChrono acquisition archive.
The additional recordings from Sites~2 and 3 contribute 15 sessions and 171,185 samples, bringing the collection to 175 sessions and 730,760 samples across three field sites. The total recording duration is 51,046~s, with a data volume of approximately 18~TB.

\begin{table}[t]
\centering
\caption{
    \textbf{Multi-site acquisition statistics of AgriChrono.}
    Recording sessions, duration, sample counts, and data volume across the three field sites.
}
\label{tab:supp_full_dataset_stats}
\small
\renewcommand{\arraystretch}{1.2}
\setlength{\tabcolsep}{10pt}
\adjustbox{max width=\columnwidth}{
    \begin{tabular}{lrrrr}
        \toprule
        \textbf{Site} & \textbf{Sessions} & \textbf{Duration (s)} & \textbf{Samples} & \textbf{Size} \\
        \midrule
        Site~1 & 160 & 37,932 & 559,575 & 13.7\,TB \\
        Site~2 & 7 & 10,235 & 129,088 & 3.2\,TB \\
        Site~3 & 8 & 2,879 & 42,097 & 1.1\,TB \\
        \midrule
        \textbf{Total} & \textbf{175} & \textbf{51,046} & \textbf{730,760} & \textbf{18\,TB} \\
        \bottomrule
    \end{tabular}
}
\end{table}

%%%%%%%%%%%%%%%%%%%%%%%%%%%%%%%%%%%%%%%%%%%%

\subsection{Camera Pose Trajectory Analysis}
\label{supp:pose_analysis}

\begin{table*}[pos=h]
    \centering
    \caption{
        Inter-frame displacement statistics for benchmark dataset. Mean ($\mu$) is compared against 0.09\,m, while lower standard deviation ($s$) and coefficient of variation ($CV=s/\mu$) indicate less displacement variability. Bold indicates the closest mean or lowest variability in each scenario.
    }
    \label{tab:supp_pose_analysis}
    \small
    \renewcommand{\arraystretch}{1.15}
    \adjustbox{max width=\textwidth}{
        \begin{tabular}{llccccccc}
            \toprule
            & &
            \multicolumn{4}{c}{\textbf{Lighting Variance (Day 19)}} &
            \multicolumn{3}{c}{\textbf{Growth Span (06:00)}} \\
            \cmidrule(lr){3-6}
            \cmidrule(l){7-9}
            \textbf{Method} & \textbf{Metric} &
            06:00 & 11:00 & 16:00 & 21:00 &
            Day 6 & Day 13 & Day 20 \\
            \midrule
            \multirow{3}{*}{VIO}
            & Mean ($\mu$, $\rightarrow 0.09$)
            & \textbf{0.0920} & \textbf{0.0912} & \textbf{0.0909} & \textbf{0.0915}
            & 0.0887 & \textbf{0.0882} & \textbf{0.0912} \\
            & Std. Dev. ($s\downarrow$)
            & \textbf{0.0131} & \textbf{0.0063} & 0.0049 & \textbf{0.0061}
            & 0.0257 & 0.0108 & 0.0074 \\
            & CV ($s/\mu\downarrow$)
            & \textbf{0.1418} & \textbf{0.0688} & \textbf{0.0542} & \textbf{0.0662}
            & 0.2892 & \textbf{0.1228} & \textbf{0.0809} \\
            \midrule
            \multirow{3}{*}{VIO+BA}
            & Mean ($\mu$, $\rightarrow 0.09$)
            & 0.0934 & 0.0914 & 0.0875 & \textbf{0.0915}
            & \textbf{0.0891} & 0.1006 & 0.0913 \\
            & Std. Dev. ($s\downarrow$)
            & 0.0263 & 0.0088 & \textbf{0.0048} & 0.0075
            & 0.0263 & 0.0186 & 0.0093 \\
            & CV ($s/\mu\downarrow$)
            & 0.2815 & 0.0958 & 0.0551 & 0.0816
            & 0.2952 & 0.1849 & 0.1015 \\
            \midrule
            \multirow{3}{*}{VGGT}
            & Mean ($\mu$, $\rightarrow 0.09$)
            & 0.0416 & 0.0486 & 0.0386 & 0.0281
            & 0.0286 & 0.0239 & 0.0249 \\
            & Std. Dev. ($s\downarrow$)
            & 0.0608 & 0.0887 & 0.0367 & 0.0216
            & \textbf{0.0079} & \textbf{0.0092} & \textbf{0.0045} \\
            & CV ($s/\mu\downarrow$)
            & 1.4618 & 1.8267 & 0.9520 & 0.7700
            & \textbf{0.2775} & 0.3825 & 0.1823 \\
            \bottomrule
        \end{tabular}
    }
\end{table*}

Camera-pose trajectories were examined using seven Site~1 recordings corresponding to lighting-variance and growth-span observations. The comparison includes recorded visual--inertial odometry (VIO) trajectories, trajectories refined using bundle adjustment (VIO+BA), and poses estimated using VGGT~\cite{wang2025vggt}. Inter-frame displacement was calculated as $d_i = \|\mathbf{t}_{i+1} - \mathbf{t}_i\|_2$, where $\mathbf{t}_i$ denotes camera translation at frame $i$. The displacement distribution was summarized using its mean ($\mu$), standard deviation ($s$), and coefficient of variation ($CV=s/\mu$). The analysis uses image sequences obtained by sampling every seventh frame from recordings. Table~\ref{tab:supp_pose_analysis} presents the results for four acquisition times on Day~19 and three observation dates at 06:00.

Across the seven sequences, VIO produces mean inter-frame displacements close to 0.09~m, with CV values ranging from 0.0542 to 0.2892. The original VIO trajectories exhibit lower CV values than VIO+BA in all seven sequences and lower values than VGGT in six of the seven sequences. The COLMAP-based structure-from-motion experiment reported in the original AgriChrono analysis did not produce usable camera trajectories for the selected field recordings~\cite{schoenberger2016sfm}. These statistics characterize local displacement consistency rather than absolute pose accuracy, as ground-truth camera trajectories are unavailable.

%% file: supplementary/sections/S03_nvs_benchmark.tex
\section{NVS Benchmark Evaluation Details}
\label{supp:benchmark}

\subsection{Canola Region Mask Construction}
\label{supp:canola_mask}

\begin{figure}[pos=h]
    \centering
    \includegraphics[width=\textwidth]{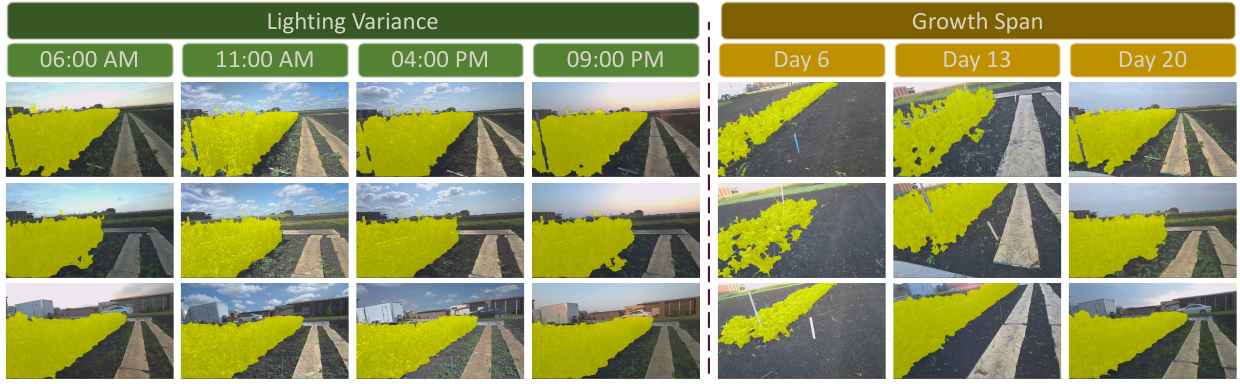}
    \caption{
        \textbf{Visualization of Canola Region annotations.}
        Representative reference RGB images and their corresponding manually annotated Canola Region masks. The same masks are used to evaluate rendering quality within the Canola Region across all methods.
    }
    \label{fig:supp_canola_mask}
\end{figure}

To evaluate rendering quality within the Canola Region, we construct Canola Region masks for all 350 test images across the seven AgriChrono benchmark scenarios, comprising 50 images per scenario. The masks are manually annotated with the assistance of SAM 3~\citep{carion2026sam3}, using the reference RGB images. Binary evaluation masks are obtained by identifying pixels marked with the annotation color (250, 250, 55), assigning one to the annotated Canola Region and zero to the remaining pixels.
The same reference-derived masks are applied to the rendered images from all evaluated methods, ensuring consistent spatial regions for comparison. These masks are used exclusively for evaluation and are not involved in model training. This region-specific evaluation focuses on the reproduction of crop appearance and fine image details, including thin stems, overlapping leaves, and partially occluded structures, complementing Whole Image evaluation. Figure~\ref{fig:supp_canola_mask} illustrates representative reference images and their corresponding Canola Region masks.

\subsection{Canola Region Detail Metrics}
\label{supp:canola_metrics}

In addition to SSIM and LPIPS, we employ the Gradient-Energy Ratio (GER) and Edge F-score to characterize fine-detail reproduction within the Canola Region. These metrics quantify gradient-energy magnitude and spatial edge correspondence, respectively. Regional SSIM is computed using the NerfBaselines implementation with \texttt{reduce=False}. The resulting SSIM maps are averaged across RGB channels and pooled over all valid Canola Region pixels across test images, excluding a five-pixel image border. Regional LPIPS is computed using AlexNet-based spatial LPIPS (\texttt{net='alex', spatial=True}) with RGB inputs normalized to $[-1,1]$. The resulting $540\times960$ spatial LPIPS maps are directly masked at the same resolution without resizing. For each scenario, LPIPS values are pooled over all Canola Region pixels across test images, rather than averaging per-frame scores. Canola pixels are identified by the yellow $(250,250,55)$ label in manually annotated masks, and frames with empty masks are excluded. The same metric implementations and reference-derived masks are used for all baseline methods and AgriSplat.

\myparagraph{Gradient-Energy Ratio (GER)}
Motivated by gradient-based image quality assessment~\citep{xue2014gradient}, GER compares gradient-energy magnitudes between rendered and reference images within the Canola Region. Let $I_i$ and $\hat{I}_i$ denote the reference and rendered images for test frame $i$, respectively, and let $M_i$ denote the corresponding binary Canola Region mask. RGB images are converted to normalized BT.601 luminance, $Y(I)=0.299R+0.587G+0.114B$, with values in $[0,1]$. The gradient energy is defined as
\begin{equation}
    E(I,M) =
    \sum_{i=1}^{N}
    \sum_{\mathbf{p}\in\Omega_i}
    M_i(\mathbf{p})
    \left\|
    \nabla_S Y(I_i)(\mathbf{p})
    \right\|_2^2,
\end{equation}
where $N=50$ is the total number of test images per scenario, $\Omega_i$ denotes the image domain, and $\nabla_S$ represents the unnormalized $3\times3$ Sobel gradient. Gradients are computed over the full images before applying the masks. The Canola Region gradient-energy ratio is then calculated as
\begin{equation}
    \mathrm{GER}_{C} =
    \frac{E(\hat{I},M)}{E(I,M)}.
\end{equation}
Frames with empty masks contribute zero to the accumulated energies. Values closer to one indicate greater agreement in gradient-energy magnitude between rendered and reference images, while values below or above one indicate lower or higher gradient energy in the rendered Canola Region, respectively.

\myparagraph{Edge F-score}
Edge F-score evaluates spatial edge correspondence between rendered and reference images within the Canola Region. For each image, normalized BT.601 luminance is converted to 8-bit intensity using $\operatorname{round}(255Y)$, followed by Gaussian smoothing with $\sigma=1$. Edges are extracted using the Canny detector~\citep{canny1986computational} with lower and upper thresholds of 30 and 90, respectively, and the L2 gradient norm. The resulting reference and rendered edge maps are independently restricted to the Canola Region before matching. Following precision-recall-based boundary evaluation~\citep{arbelaez2011contour}, edge correspondence is determined using the exact Euclidean distance transform with a tolerance of $\delta=2$ pixels.

Let $\mathcal{E}_i$ and $\hat{\mathcal{E}}_i$ denote the masked reference and rendered edge sets for test frame $i$, respectively. Precision and recall are defined as
\begin{equation}
    \begin{aligned}
    P_C &=
    \frac{
        \displaystyle\sum_{i=1}^{N}
        \sum_{\hat{\mathbf{p}}\in\hat{\mathcal{E}}_i}
        \mathbb{1}\left[
            d(\hat{\mathbf{p}},\mathcal{E}_i)
            \leq\delta
        \right]
    }{
        \displaystyle\sum_{i=1}^{N}
        |\hat{\mathcal{E}}_i|
    }, \qquad
    R_C &=
    \frac{
        \displaystyle\sum_{i=1}^{N}
        \sum_{\mathbf{p}\in\mathcal{E}_i}
        \mathbb{1}\left[
            d(\mathbf{p},\hat{\mathcal{E}}_i)
            \leq\delta
        \right]
    }{
        \displaystyle\sum_{i=1}^{N}
        |\mathcal{E}_i|
    },
    \end{aligned}
\end{equation}
where $d(\mathbf{p},\mathcal{E}_i)$ denotes the Euclidean distance from pixel $\mathbf{p}$ to the nearest pixel in the masked edge set $\mathcal{E}_i$, and $\mathbb{1}[\cdot]$ is the indicator function. The Edge F-score is then computed as
\begin{equation}
    \mathrm{EdgeF}_{C} =
    \frac{2P_CR_C}{P_C+R_C}.
\end{equation}
Precision and recall are obtained by pooling matched and total edge-pixel counts across the test images within each scenario. Frames with empty masks do not contribute to these counts. Higher Edge F-scores indicate greater spatial correspondence between rendered and reference image edges.

\clearpage

\begin{table*}[t]
    \centering
    \caption{
        \textbf{Full scenario-wise Canola Region image similarity.}
        Higher SSIM and lower LPIPS indicate better performance. \textbf{Bold} and \underline{underlined} denote the best and second-best.
        Results are reported across seven independent AgriChrono scenarios.
        Day 19 captures time-of-day variation, while Days 6, 13, and 20 were recorded at 06:00.
    }
    \label{tab:supp_canola_similarity}
    \renewcommand{\arraystretch}{1.1}
    \setlength\tabcolsep{1.5pt}
    \small
    \adjustbox{max width=\columnwidth}{%
    \begin{tabular}{l|cc|cc|cc|cc|cc|cc|cc}
        \toprule
        \multirow{3}{*}{\textbf{Method}}
        & \multicolumn{8}{c|}{\textbf{Lighting Variance (Day 19)}}
        & \multicolumn{6}{c}{\textbf{Growth Span (06:00)}} \\
        \cmidrule(lr){2-9}\cmidrule(lr){10-15}
        & \multicolumn{2}{c|}{\textbf{06:00}} & \multicolumn{2}{c|}{\textbf{11:00}} & \multicolumn{2}{c|}{\textbf{16:00}} & \multicolumn{2}{c|}{\textbf{21:00}}
        & \multicolumn{2}{c|}{\textbf{Day 6}} & \multicolumn{2}{c|}{\textbf{Day 13}} & \multicolumn{2}{c}{\textbf{Day 20}} \\
        \cmidrule(lr){2-3}\cmidrule(lr){4-5}\cmidrule(lr){6-7}\cmidrule(lr){8-9}\cmidrule(lr){10-11}\cmidrule(lr){12-13}\cmidrule(lr){14-15}
        & SSIM$_C$ & LPIPS$_C$ & SSIM$_C$ & LPIPS$_C$ & SSIM$_C$ & LPIPS$_C$ & SSIM$_C$ & LPIPS$_C$ & SSIM$_C$ & LPIPS$_C$ & SSIM$_C$ & LPIPS$_C$ & SSIM$_C$ & LPIPS$_C$ \\
        \midrule
        SeaThru-NeRF
        & 0.457 & 0.849 & 0.204 & 0.984 & 0.278 & 0.914 & 0.470 & 0.828 & 0.183 & 0.751 & 0.386 & 0.753 & 0.374 & 0.872 \\
        \midrule
        Nerfacto
        & 0.460 & 0.782 & 0.199 & 0.853 & 0.283 & 0.791 & 0.471 & 0.780 & 0.200 & 0.645 & 0.376 & 0.730 & 0.373 & 0.791 \\
        \midrule
        Tetra-NeRF
        & 0.506 & 0.670 & 0.287 & 0.764 & 0.353 & 0.708 & 0.523 & 0.653 & 0.414 & 0.431 & 0.501 & 0.534 & 0.444 & 0.684 \\
        \midrule
        Zip-NeRF
        & 0.553 & 0.482 & \textbf{0.448} & \underline{0.403} & \textbf{0.468} & 0.410 & 0.560 & 0.474 & \underline{0.620} & \underline{0.238} & \underline{0.669} & \underline{0.282} & \textbf{0.602} & \underline{0.336} \\
        \midrule
        3DGS
        & 0.558 & 0.582 & 0.361 & 0.618 & 0.412 & 0.611 & 0.562 & 0.593 & 0.562 & 0.300 & 0.596 & 0.403 & 0.498 & 0.606 \\
        \midrule
        Scaffold-GS
        & 0.512 & 0.593 & 0.318 & 0.606 & 0.392 & 0.566 & 0.532 & 0.587 & 0.516 & 0.323 & 0.554 & 0.437 & 0.447 & 0.595 \\
        \midrule
        Taming 3DGS
        & 0.564 & 0.543 & 0.367 & 0.587 & 0.412 & 0.583 & 0.562 & 0.565 & 0.573 & 0.309 & 0.605 & 0.397 & 0.507 & 0.564 \\
        \midrule
        3DGRT
        & 0.547 & 0.570 & 0.352 & 0.592 & 0.403 & 0.584 & 0.551 & 0.581 & 0.546 & 0.297 & 0.588 & 0.408 & 0.490 & 0.579 \\
        \midrule
        H3DGS
        & 0.574 & \underline{0.377} & 0.366 & 0.435 & 0.414 & \underline{0.406} & 0.574 & \underline{0.393} & 0.547 & 0.253 & 0.615 & 0.300 & 0.523 & 0.376 \\
        \midrule
        GS-W
        & 0.466 & 0.807 & 0.213 & 0.945 & 0.317 & 0.855 & 0.495 & 0.777 & 0.308 & 0.650 & 0.461 & 0.662 & 0.361 & 0.840 \\
        \midrule
        3D-MCMC
        & 0.481 & 0.817 & 0.268 & 0.853 & 0.334 & 0.822 & 0.492 & 0.833 & 0.470 & 0.439 & 0.508 & 0.579 & 0.419 & 0.805 \\
        \midrule
        PGSR
        & 0.527 & 0.647 & 0.334 & 0.661 & 0.387 & 0.670 & 0.532 & 0.674 & 0.479 & 0.372 & 0.526 & 0.527 & 0.472 & 0.635 \\
        \midrule
        gsplat
        & 0.546 & 0.604 & 0.349 & 0.635 & 0.402 & 0.628 & 0.554 & 0.607 & 0.543 & 0.314 & 0.580 & 0.429 & 0.490 & 0.619 \\
        \midrule
        Octree-GS
        & \underline{0.589} & 0.393 & 0.390 & 0.469 & 0.429 & 0.435 & \underline{0.581} & 0.428 & 0.590 & 0.268 & 0.625 & 0.325 & 0.528 & 0.419 \\
        \midrule
        3DGUT
        & 0.475 & 0.737 & 0.249 & 0.790 & 0.315 & 0.754 & 0.487 & 0.747 & 0.318 & 0.509 & 0.435 & 0.611 & 0.405 & 0.743 \\
        \midrule
        \textbf{AgriSplat}
        & \textbf{0.609} & \textbf{0.277} & \underline{0.411} & \textbf{0.325} & \underline{0.443} & \textbf{0.327} & \textbf{0.596} & \textbf{0.291} & \textbf{0.634} & \textbf{0.186} & \textbf{0.670} & \textbf{0.213} & \underline{0.556} & \textbf{0.288} \\
        \bottomrule
    \end{tabular}}
\end{table*}

\subsection{Detailed Scenario-wise Results}
\label{supp:benchmark_metrics}

This section provides scenario-wise quantitative results for the 15 baseline methods and AgriSplat, complementing the aggregate and representative comparisons presented in Section 6 of the paper. The evaluation comprises four Lighting Variance scenarios acquired at 06:00, 11:00, 16:00, and 21:00 on Day 19, and three Growth Span scenarios acquired at 06:00 on Days 6, 13, and 20. Each scenario is trained and evaluated independently, with the aggregate results obtained by equally weighting the seven scenario-level measurements. Tables~\ref{tab:supp_canola_similarity} and \ref{tab:supp_canola_edges} report Canola Region SSIM, LPIPS, GER, and Edge F-score, while Table~\ref{tab:supp_whole} provides Whole Image PSNR, SSIM, and LPIPS.

Across the Lighting Variance scenarios, all 15 baselines achieve their lowest Canola Region SSIM at 11:00 among the four acquisition times. However, Edge F-score exhibits a different pattern. For example, Octree-GS achieves SSIM$_C$ values of 0.390 at 11:00 and 0.581 at 21:00, whereas its EdgeF$_C$ values are 0.819 and 0.511, respectively. A similar difference is observed for AgriSplat, which achieves higher SSIM$_C$ at 21:00 but higher EdgeF$_C$ at 11:00. These results indicate that image similarity and edge correspondence do not necessarily follow the same trends across acquisition conditions.

Across the Growth Span scenarios, all evaluated methods achieve their highest Canola Region SSIM on Day 13, whereas their highest Edge F-scores occur on Day 6. For example, Zip-NeRF achieves SSIM$_C$ values of 0.620, 0.669, and 0.602 on Days 6, 13, and 20, respectively, while its EdgeF$_C$ decreases from 0.890 to 0.712 and 0.618. These differences suggest that conventional image similarity and boundary correspondence capture different aspects of reconstruction quality across crop-development stages.

\clearpage
Comparing Whole Image and Canola Region evaluations further reveals differences in the relative performance of individual methods. On Day 6, Octree-GS achieves higher Whole Image PSNR and SSIM than Zip-NeRF (26.07 dB versus 25.94 dB and 0.700 versus 0.610), whereas Zip-NeRF achieves higher Canola Region SSIM (0.620 versus 0.590) and lower Canola Region LPIPS (0.238 versus 0.268). Across all seven scenarios, AgriSplat achieves the lowest Canola Region LPIPS, the GER closest to one, and the highest Edge F-score among the evaluated methods. These results reinforce the importance of region-specific evaluation alongside conventional Whole Image metrics.

\begin{table*}[t]
    \centering
    \caption{
        \textbf{Full scenario-wise Canola Region gradient and edge evaluation.}
        GER values closer to 1 and higher Edge F-scores indicate better performance. \textbf{Bold} and \underline{underlined} denote the best and second-best.
        Results are reported across seven independent AgriChrono scenarios.
    }
    \label{tab:supp_canola_edges}
    \renewcommand{\arraystretch}{1.5}
    \setlength\tabcolsep{1.5pt}
    \small
    \adjustbox{max width=\textwidth}{%
    \begin{tabular}{l|cc|cc|cc|cc|cc|cc|cc}
        \toprule
        \multirow{3}{*}{\textbf{Method}}
        & \multicolumn{8}{c|}{\textbf{Lighting Variance (Day 19)}}
        & \multicolumn{6}{c}{\textbf{Growth Span (06:00)}} \\
        \cmidrule(lr){2-9}\cmidrule(lr){10-15}
        & \multicolumn{2}{c|}{\textbf{06:00}} & \multicolumn{2}{c|}{\textbf{11:00}} & \multicolumn{2}{c|}{\textbf{16:00}} & \multicolumn{2}{c|}{\textbf{21:00}}
        & \multicolumn{2}{c|}{\textbf{Day 6}} & \multicolumn{2}{c|}{\textbf{Day 13}} & \multicolumn{2}{c}{\textbf{Day 20}} \\
        \cmidrule(lr){2-3}\cmidrule(lr){4-5}\cmidrule(lr){6-7}\cmidrule(lr){8-9}\cmidrule(lr){10-11}\cmidrule(lr){12-13}\cmidrule(lr){14-15}
        & GER$_C$ & EdgeF$_C$ & GER$_C$ & EdgeF$_C$ & GER$_C$ & EdgeF$_C$ & GER$_C$ & EdgeF$_C$ & GER$_C$ & EdgeF$_C$ & GER$_C$ & EdgeF$_C$ & GER$_C$ & EdgeF$_C$ \\
        \midrule
        SeaThru-NeRF
        & 0.023 & 0.020 & 0.014 & 0.014 & 0.020 & 0.020 & 0.027 & 0.015 & 0.034 & 0.061 & 0.022 & 0.006 & 0.023 & 0.013 \\
        \midrule
        Nerfacto
        & 0.040 & 0.042 & 0.035 & 0.056 & 0.045 & 0.102 & 0.030 & 0.045 & 0.104 & 0.403 & 0.038 & 0.017 & 0.039 & 0.032 \\
        \midrule
        Tetra-NeRF
        & 0.070 & 0.048 & 0.089 & 0.266 & 0.096 & 0.271 & 0.078 & 0.062 & 0.261 & 0.711 & 0.161 & 0.325 & 0.086 & 0.095 \\
        \midrule
        Zip-NeRF
        & 0.253 & 0.369 & 0.367 & 0.763 & 0.362 & 0.712 & 0.244 & 0.391 & 0.555 & 0.890 & 0.454 & 0.712 & 0.412 & 0.618 \\
        \midrule
        3DGS
        & 0.227 & 0.291 & 0.311 & 0.660 & 0.301 & 0.604 & 0.223 & 0.290 & 0.615 & 0.883 & 0.452 & 0.667 & 0.249 & 0.388 \\
        \midrule
        Scaffold-GS
        & 0.290 & 0.328 & 0.433 & 0.771 & 0.345 & 0.676 & 0.289 & 0.317 & 0.554 & 0.870 & 0.468 & 0.648 & 0.378 & 0.492 \\
        \midrule
        Taming 3DGS
        & 0.253 & 0.334 & 0.324 & 0.702 & 0.308 & 0.632 & 0.232 & 0.290 & 0.602 & 0.886 & 0.459 & 0.675 & 0.285 & 0.444 \\
        \midrule
        3DGRT
        & 0.236 & 0.310 & 0.315 & 0.679 & 0.311 & 0.626 & 0.223 & 0.291 & 0.595 & 0.882 & 0.423 & 0.657 & 0.256 & 0.408 \\
        \midrule
        H3DGS
        & 0.440 & 0.525 & \underline{0.502} & \underline{0.835} & \underline{0.511} & 0.794 & \underline{0.444} & \underline{0.521} & 0.659 & 0.901 & \underline{0.562} & 0.736 & \underline{0.497} & \underline{0.650} \\
        \midrule
        GS-W
        & 0.058 & 0.035 & 0.080 & 0.221 & 0.079 & 0.210 & 0.057 & 0.048 & 0.124 & 0.453 & 0.081 & 0.145 & 0.130 & 0.101 \\
        \midrule
        3D-MCMC
        & 0.049 & 0.064 & 0.103 & 0.314 & 0.104 & 0.307 & 0.043 & 0.073 & 0.431 & 0.806 & 0.210 & 0.416 & 0.071 & 0.116 \\
        \midrule
        PGSR
        & 0.179 & 0.229 & 0.261 & 0.588 & 0.249 & 0.532 & 0.164 & 0.234 & 0.513 & 0.822 & 0.294 & 0.492 & 0.208 & 0.340 \\
        \midrule
        gsplat
        & 0.196 & 0.255 & 0.293 & 0.640 & 0.279 & 0.581 & 0.204 & 0.261 & 0.603 & 0.877 & 0.411 & 0.631 & 0.226 & 0.350 \\
        \midrule
        Octree-GS
        & \underline{0.467} & \underline{0.564} & 0.449 & 0.819 & 0.502 & \underline{0.799} & 0.419 & 0.511 & \underline{0.674} & \underline{0.905} & 0.558 & \underline{0.740} & 0.472 & 0.646 \\
        \midrule
        3DGUT
        & 0.058 & 0.063 & 0.126 & 0.352 & 0.114 & 0.313 & 0.049 & 0.057 & 0.305 & 0.722 & 0.126 & 0.221 & 0.079 & 0.097 \\
        \midrule
        \textbf{AgriSplat}
        & \textbf{0.785} & \textbf{0.686} & \textbf{0.824} & \textbf{0.888} & \textbf{0.798} & \textbf{0.844} & \textbf{0.793} & \textbf{0.680} & \textbf{0.977} & \textbf{0.924} & \textbf{0.916} & \textbf{0.807} & \textbf{0.838} & \textbf{0.751} \\
        \bottomrule
    \end{tabular}}
\end{table*}

\begin{table*}[t]
    \centering
    \caption{
        \textbf{Full scenario-wise Whole Image evaluation.}
        PSNR and SSIM are higher-is-better metrics, whereas LPIPS is lower-is-better. Results are reported for seven independent scenarios. \textbf{Bold} and \underline{underlined} denote the best and second-best.
    }
    \label{tab:supp_whole}
    \renewcommand{\arraystretch}{1.0}
    \setlength\tabcolsep{8pt}
    \small
    \adjustbox{max width=\textwidth}{%
    \begin{tabular}{ll|rrrr|rrr}
        \toprule
        & & \multicolumn{4}{c|}{\textit{Lighting Variance (Day 19)}}
            & \multicolumn{3}{c}{\textit{Growth Span (06:00)}} \\
        \cmidrule(lr){3-6}\cmidrule(lr){7-9}
        \textbf{Method} & \textbf{Metric}
        & 06:00 & 11:00 & 16:00 & 21:00 & Day 6 & Day 13 & Day 20 \\
        \midrule
        \multirow{3}{*}{SeaThru-NeRF~\citep{levy2023seathru}}
        & PSNR$_W$ & 25.37 & 21.42 & 22.88 & 26.46 & 22.69 & 21.49 & 24.08 \\
        & SSIM$_W$ & 0.627 & 0.441 & 0.498 & 0.653 & 0.443 & 0.460 & 0.568 \\
        & LPIPS$_W$ & 0.627 & 0.705 & 0.681 & 0.608 & 0.713 & 0.716 & 0.646 \\
        \midrule
        \multirow{3}{*}{Nerfacto~\citep{nerfstudio}}
        & PSNR$_W$ & 23.70 & 20.27 & 21.57 & 24.74 & 22.13 & 18.81 & 23.19 \\
        & SSIM$_W$ & 0.639 & 0.435 & 0.501 & 0.655 & 0.445 & 0.445 & 0.581 \\
        & LPIPS$_W$ & 0.489 & 0.537 & 0.498 & 0.464 & 0.638 & 0.691 & 0.488 \\
        \midrule
        \multirow{3}{*}{Tetra-NeRF~\citep{kulhanek2023tetranerf}}
        & PSNR$_W$ & 28.23 & 24.06 & 25.56 & 29.34 & 25.32 & 27.10 & 27.39 \\
        & SSIM$_W$ & 0.733 & 0.607 & 0.646 & 0.756 & 0.598 & 0.658 & 0.702 \\
        & LPIPS$_W$ & 0.429 & 0.492 & 0.477 & 0.391 & 0.402 & 0.393 & 0.424 \\
        \midrule
        \multirow{3}{*}{Zip-NeRF~\citep{barron2023zipnerf}}
        & PSNR$_W$ & \textbf{29.58} & \textbf{25.92} & \textbf{27.47} & \textbf{30.67} & 25.94 & \textbf{28.84} & \textbf{29.42} \\
        & SSIM$_W$ & 0.795 & \textbf{0.756} & \textbf{0.766} & 0.807 & 0.610 & \underline{0.774} & \textbf{0.810} \\
        & LPIPS$_W$ & 0.252 & \textbf{0.200} & \textbf{0.203} & \underline{0.237} & 0.417 & \underline{0.216} & \underline{0.194} \\
        \midrule
        \multirow{3}{*}{3DGS~\citep{kerbl20233d}}
        & PSNR$_W$ & 27.05 & 23.98 & 25.02 & 28.72 & 25.45 & 26.84 & 27.11 \\
        & SSIM$_W$ & 0.768 & 0.669 & 0.694 & 0.784 & 0.642 & 0.706 & 0.742 \\
        & LPIPS$_W$ & 0.364 & 0.381 & 0.384 & 0.339 & 0.392 & 0.361 & 0.361 \\
        \midrule
        \multirow{3}{*}{Scaffold-GS~\citep{scaffoldgs}}
        & PSNR$_W$ & 25.62 & 20.97 & 24.78 & 26.69 & 24.85 & 25.15 & 24.18 \\
        & SSIM$_W$ & 0.734 & 0.603 & 0.678 & 0.761 & 0.637 & 0.685 & 0.692 \\
        & LPIPS$_W$ & 0.375 & 0.390 & 0.369 & 0.342 & 0.351 & 0.365 & 0.375 \\
        \midrule
        \multirow{3}{*}{Taming 3DGS~\citep{Taming3DGS}}
        & PSNR$_W$ & 28.00 & 24.12 & 25.38 & 29.05 & 25.73 & 27.24 & 27.39 \\
        & SSIM$_W$ & 0.777 & 0.677 & 0.699 & 0.786 & 0.650 & 0.715 & 0.747 \\
        & LPIPS$_W$ & 0.341 & 0.373 & 0.373 & 0.335 & 0.391 & 0.355 & 0.349 \\
        \midrule
        \multirow{3}{*}{3DGRT~\citep{3dgrt2024}}
        & PSNR$_W$ & 26.70 & 23.60 & 24.73 & 28.46 & 25.32 & 26.42 & 26.30 \\
        & SSIM$_W$ & 0.765 & 0.662 & 0.690 & 0.783 & 0.635 & 0.699 & 0.737 \\
        & LPIPS$_W$ & 0.348 & 0.371 & 0.375 & 0.321 & 0.385 & 0.351 & 0.346 \\
        \midrule
        \multirow{3}{*}{H3DGS~\citep{hierarchicalgaussians24}}
        & PSNR$_W$ & 27.79 & \underline{24.28} & \underline{25.70} & 29.30 & 25.48 & 27.31 & 27.30 \\
        & SSIM$_W$ & 0.792 & 0.697 & 0.723 & 0.804 & 0.666 & 0.739 & 0.767 \\
        & LPIPS$_W$ & 0.255 & 0.275 & 0.261 & 0.239 & 0.293 & 0.249 & 0.247 \\
        \midrule
        \multirow{3}{*}{GS-W~\citep{zhang2024GS-W}}
        & PSNR$_W$ & 20.78 & 16.41 & 20.54 & 22.80 & 23.44 & 23.47 & 19.75 \\
        & SSIM$_W$ & 0.654 & 0.477 & 0.564 & 0.698 & 0.512 & 0.581 & 0.585 \\
        & LPIPS$_W$ & 0.563 & 0.636 & 0.587 & 0.497 & 0.609 & 0.555 & 0.574 \\
        \midrule
        \multirow{3}{*}{3D-MCMC~\citep{kheradmand20243d}}
        & PSNR$_W$ & 27.03 & 23.31 & 24.55 & 28.28 & 24.46 & 25.96 & 26.34 \\
        & SSIM$_W$ & 0.715 & 0.581 & 0.621 & 0.734 & 0.548 & 0.625 & 0.677 \\
        & LPIPS$_W$ & 0.505 & 0.550 & 0.536 & 0.476 & 0.611 & 0.518 & 0.496 \\
        \midrule
        \multirow{3}{*}{PGSR~\citep{chen2024pgsr}}
        & PSNR$_W$ & 26.89 & 23.58 & 24.72 & 28.32 & 24.28 & 25.56 & 26.57 \\
        & SSIM$_W$ & 0.744 & 0.637 & 0.664 & 0.759 & 0.549 & 0.629 & 0.712 \\
        & LPIPS$_W$ & 0.412 & 0.426 & 0.436 & 0.393 & 0.601 & 0.494 & 0.407 \\
        \midrule
        \multirow{3}{*}{gsplat~\citep{ye2025gsplat}}
        & PSNR$_W$ & 27.26 & 23.86 & 24.92 & 28.80 & 25.30 & 26.72 & 27.08 \\
        & SSIM$_W$ & 0.766 & 0.658 & 0.685 & 0.783 & 0.629 & 0.694 & 0.738 \\
        & LPIPS$_W$ & 0.370 & 0.395 & 0.401 & 0.342 & 0.410 & 0.381 & 0.367 \\
        \midrule
        \multirow{3}{*}{Octree-GS~\citep{ren2024octree}}
        & PSNR$_W$ & 28.49 & 24.25 & 25.52 & 29.56 & \underline{26.07} & 27.60 & 27.70 \\
        & SSIM$_W$ & \underline{0.803} & 0.700 & 0.721 & \underline{0.808} & \underline{0.700} & 0.746 & 0.774 \\
        & LPIPS$_W$ & \underline{0.251} & 0.313 & 0.300 & 0.251 & \underline{0.281} & 0.274 & 0.264 \\
        \midrule
        \multirow{3}{*}{3DGUT~\citep{wu20253dgut}}
        & PSNR$_W$ & 26.51 & 22.96 & 24.12 & 27.72 & 23.71 & 24.85 & 25.87 \\
        & SSIM$_W$ & 0.693 & 0.552 & 0.587 & 0.713 & 0.512 & 0.559 & 0.648 \\
        & LPIPS$_W$ & 0.469 & 0.502 & 0.501 & 0.448 & 0.560 & 0.527 & 0.466 \\
        \midrule
        \multirow{3}{*}{\textbf{AgriSplat (Ours)}}
        & PSNR$_W$ & \underline{28.69} & \underline{24.28} & 25.64 & \underline{29.81} & \textbf{26.39} & \underline{28.02} & \underline{27.83} \\
        & SSIM$_W$ & \textbf{0.823} & \underline{0.730} & \underline{0.745} & \textbf{0.827} & \textbf{0.740} & \textbf{0.787} & \underline{0.799} \\
        & LPIPS$_W$ & \textbf{0.177} & \underline{0.223} & \underline{0.223} & \textbf{0.165} & \textbf{0.181} & \textbf{0.163} & \textbf{0.180} \\
        \bottomrule
    \end{tabular}}
\end{table*}

%% file: supplementary/sections/S04_acquisition_quality.tex
\clearpage
\section{Additional Ablation and Hyperparameter Analyses}
\label{supp:ablation}

\subsection{Qualitative Ablation Comparisons}
\label{supp:ablation_qualitative}

\begin{figure*}[pos=h]
    \centering
    \includegraphics[width=\textwidth]{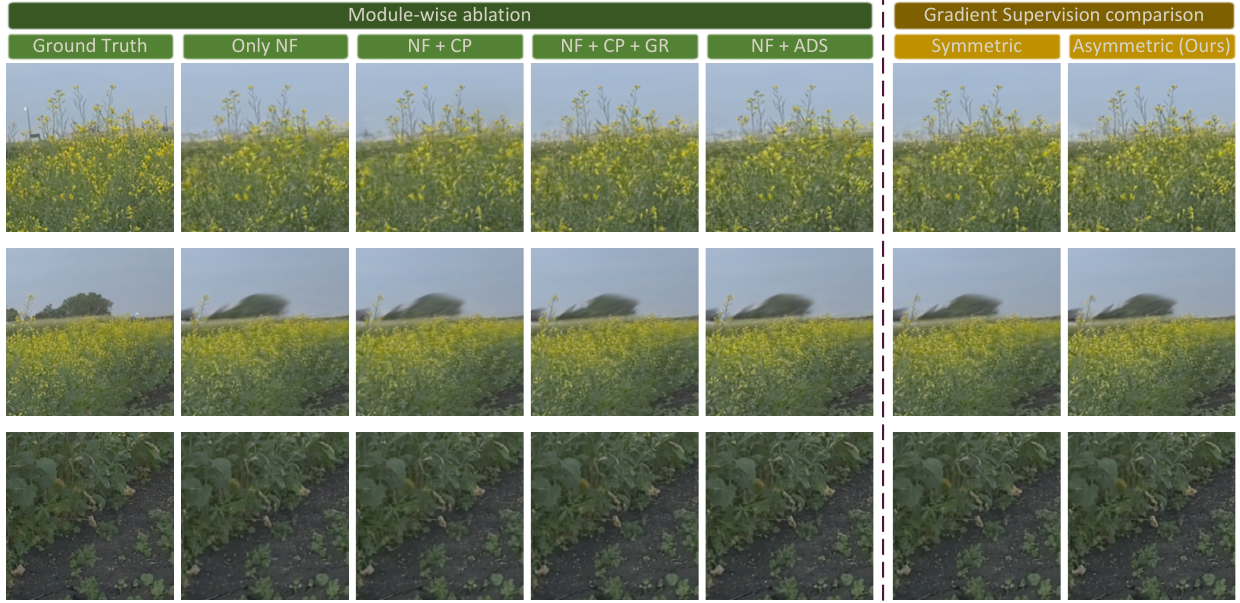}
    \caption{
        \textbf{Qualitative ablation comparisons on the Day 20 scenario.} Three representative cropped regions are shown. \textbf{Left (Module-wise Ablation):} Ground-truth images and renderings obtained using NF, CP, GR, and ADS in the indicated combinations. \textbf{Right (Gradient Supervision Comparison):} Symmetric and asymmetric aligned gradient penalties under the same CAR configuration. The asymmetric variant corresponds to the full AgriSplat configuration (NF+CP+GR+ADS). The comparisons illustrate differences in
        vegetation boundaries, overlapping structures, and local contrast.
    }
    \label{fig:ablation_qualitative}
\end{figure*}

Figure~\ref{fig:ablation_qualitative} provides qualitative comparisons of the ablation configurations evaluated in Section 6.4 of the main paper. Three representative cropped regions from the Day 20 scenario are used to examine differences in fine-detail reproduction. Under visibility-relaxed initialization (NF), introducing Contribution Pruning (CP) produces little perceptible change in rendering quality, despite reducing the average anchor count by approximately 42\% (from 785.8k to 455.6k). 
Adding Gradient-Ranked Reinvestment (GR) improves visual quality compared with NF+CP, while subsequent Asymmetric Directional Supervision (ADS) further enhances the sharpness of vegetation boundaries and fine structures.
Under the same CAR configuration, asymmetric gradient supervision also produces more distinct local contrast than its symmetric counterpart, particularly around bright vegetation details.
These observations complement the quantitative ablation results by illustrating the respective contributions of anchor pruning, reinvestment, and directional supervision.

\subsection{Hyperparameter Selection and Sensitivity Analysis}
\label{supp:hyperparameters}

\begin{table*}[h!]
    \centering
    \caption{
        \textbf{Hyperparameter sensitivity analysis.} Mean results across three Growth Span scenarios (Days 6, 13, and 20), varying one hyperparameter at a time from the default configuration. $N$ denotes the final anchor count (thousands). \textbf{Bold} and \underline{underlined} indicate the best and second-best results within each block.
    }
    \label{tab:hyperparameters}
    \renewcommand{\arraystretch}{0.9}
    \setlength{\tabcolsep}{7pt}
    \small
    \adjustbox{max width=\textwidth}{
        \begin{tabular}{l|ccc|cccc|r}
            \toprule
            Setting & PSNR$_W\uparrow$ & SSIM$_W\uparrow$ & LPIPS$_W\downarrow$ & SSIM$_C\uparrow$ & LPIPS$_C\downarrow$ & GER$_C\rightarrow 1$ & EdgeF$_C\uparrow$ & $N$ \\
            \midrule
            \multicolumn{9}{l}{\textit{ADS weight $\lambda_{\mathrm{ADS}}$}} \\
            \midrule
            0.25 & \textbf{27.64} & \textbf{0.780} & 0.179 & \textbf{0.626} & 0.234 & 0.763 & 0.823 & 745.6 \\
            0.5 & \underline{27.57} & \underline{0.779} & \textbf{0.177} & \underline{0.624} & 0.232 & 0.814 & \underline{0.825} & 745.6 \\
            1 (ours) & 27.38 & 0.776 & \underline{0.177} & 0.618 & \underline{0.232} & \textbf{0.914} & \textbf{0.827} & 745.6 \\
            2 & 26.97 & 0.765 & 0.179 & 0.604 & \textbf{0.231} & \underline{1.113} & 0.824 & 745.6 \\
            4 & 26.41 & 0.754 & 0.193 & 0.587 & 0.243 & 1.433 & 0.810 & 745.6 \\
            \midrule
            \multicolumn{9}{l}{\textit{Contribution threshold $\theta_W$}} \\
            \midrule
            2 & 27.34 & 0.775 & \underline{0.175} & 0.615 & \underline{0.231} & \underline{0.918} & 0.826 & 867.1 \\
            4 & 27.38 & \underline{0.775} & \textbf{0.175} & 0.617 & \textbf{0.231} & \textbf{0.920} & 0.827 & 815.0 \\
            8 (ours) & \underline{27.38} & \textbf{0.776} & 0.177 & \underline{0.618} & 0.232 & 0.914 & \underline{0.827} & 745.6 \\
            16 & \textbf{27.39} & 0.775 & 0.180 & \textbf{0.620} & 0.232 & 0.908 & \textbf{0.828} & \underline{618.0} \\
            32 & 27.30 & 0.770 & 0.194 & 0.614 & 0.250 & 0.874 & 0.823 & \textbf{461.4} \\
            \midrule
            \multicolumn{9}{l}{\textit{Credit fraction}} \\
            \midrule
            0.25 & 27.27 & 0.771 & 0.191 & 0.608 & 0.252 & 0.865 & 0.820 & \textbf{593.8} \\
            0.5 & 27.25 & 0.770 & 0.188 & 0.610 & 0.248 & 0.881 & 0.821 & \underline{636.2} \\
            1 (ours) & \underline{27.38} & \underline{0.776} & \underline{0.177} & \underline{0.618} & \underline{0.232} & \underline{0.914} & \underline{0.827} & 745.6 \\
            2 & \textbf{27.47} & \textbf{0.780} & \textbf{0.161} & \textbf{0.623} & \textbf{0.213} & \textbf{0.960} & \textbf{0.832} & 937.4 \\
            \midrule
            \multicolumn{9}{l}{\textit{ADS start iteration}} \\
            \midrule
            1 & 27.36 & \underline{0.777} & \textbf{0.169} & \underline{0.619} & \textbf{0.215} & \underline{0.961} & \textbf{0.830} & 917.7 \\
            5000 & 27.35 & 0.776 & \underline{0.170} & 0.619 & \underline{0.219} & \textbf{0.962} & 0.829 & 901.5 \\
            15000 & \textbf{27.39} & \textbf{0.777} & 0.170 & \textbf{0.620} & 0.221 & 0.946 & \underline{0.829} & \underline{863.0} \\
            25000 (ours) & \underline{27.38} & 0.776 & 0.177 & 0.618 & 0.232 & 0.914 & 0.827 & \textbf{745.6} \\
            \midrule
            \multicolumn{9}{l}{\textit{Pruning interval}} \\
            \midrule
            2500 & \underline{27.37} & \textbf{0.776} & \underline{0.174} & \textbf{0.620} & \textbf{0.226} & \textbf{0.933} & \textbf{0.828} & \underline{761.3} \\
            5000 (ours) & \textbf{27.38} & 0.776 & 0.177 & \underline{0.618} & 0.232 & 0.914 & 0.827 & \textbf{745.6} \\
            10000 & 27.37 & \underline{0.776} & \textbf{0.173} & 0.616 & \underline{0.229} & \underline{0.922} & \underline{0.828} & 930.0 \\
            \midrule
            \multicolumn{9}{l}{\textit{Growth end iteration}} \\
            \midrule
            15000 & \underline{27.33} & \underline{0.773} & \underline{0.184} & \underline{0.613} & \underline{0.243} & \underline{0.888} & \underline{0.823} & \textbf{739.7} \\
            25000 (ours) & \textbf{27.38} & \textbf{0.776} & \textbf{0.177} & \textbf{0.618} & \textbf{0.232} & \textbf{0.914} & \textbf{0.827} & \underline{745.6} \\
            \bottomrule
        \end{tabular}
    }
\vspace{-5pt}
\end{table*}

Table~\ref{tab:hyperparameters} presents hyperparameter selection and sensitivity results across three Growth Span scenarios (Days 6, 13, and 20). Hyperparameters were selected through exploratory tuning using the benchmark test-view results from these scenarios. Each experiment varies one hyperparameter while keeping the remaining settings fixed. We examine the ADS weight $\lambda_{\mathrm{ADS}}$, contribution threshold $\theta_W$, credit scaling factor, ADS start iteration, pruning interval, and growth end iteration. Performance is evaluated using Whole Image and Canola Region metrics and final anchor counts.

Increasing $\lambda_{\mathrm{ADS}}$ improves crop-region gradient-energy agreement up to a certain point, but larger weights reduce image similarity and may produce excessive gradient energy. We select $\lambda_{\mathrm{ADS}}=1$ to emphasize fine-detail reproduction in canola regions while limiting the loss of overall image fidelity. The contribution threshold $\theta_W$ controls pruning aggressiveness. Higher thresholds reduce anchor counts, but excessive pruning degrades reconstruction quality. We adopt $\theta_W=8$ as a conservative balance between preserving reconstruction quality and reducing redundant anchors.

Increasing the credit scaling factor generally improves reconstruction quality but requires more anchors. Although a factor of 2 achieves better metrics, we use a factor of 1 to maintain the proposed reallocation principle, where reinvestment does not introduce more anchors than the pruning credit permits.

Earlier ADS activation improves perceptual similarity and edge correspondence but also results in more retained anchors. Applying ADS during anchor growth may affect anchor contributions and pruning decisions, potentially retaining more anchors throughout optimization. We therefore activate ADS only after the anchor topology becomes fixed at iteration 25,000, which is also the default growth end iteration of Octree-GS. This separates representation reallocation from subsequent contrast refinement.

Changing the pruning interval from 2,500 to 5,000 iterations produces only minor differences in reconstruction quality and anchor counts. However, extending the interval to 10,000 iterations substantially increases the number of retained anchors. We therefore use an interval of 5,000 iterations. Finally, extending anchor growth from 15,000 to 25,000 iterations improves reconstruction quality, supporting the default growth schedule inherited from Octree-GS.

Overall, the selected configuration prioritizes fine-detail reconstruction while maintaining controlled anchor allocation, rather than maximizing individual evaluation metrics.

%% file: cas-refs.bib
@inproceedings{zhang2018unreasonable,
  title={The unreasonable effectiveness of deep features as a perceptual metric},
  author={Zhang, Richard and Isola, Phillip and Efros, Alexei A and Shechtman, Eli and Wang, Oliver},
  booktitle={2018 IEEE/CVF conference on computer vision and pattern recognition},
  pages={586--595},
  year={2018},
  organization={IEEE}
}

@article{wang2004image,
  title={Image quality assessment: from error visibility to structural similarity},
  author={Wang, Zhou and Bovik, Alan C and Sheikh, Hamid R and Simoncelli, Eero P},
  journal={IEEE transactions on image processing},
  volume={13},
  number={4},
  pages={600--612},
  year={2004},
  publisher={IEEE}
}

@inproceedings{hore2010image,
  title={Image quality metrics: PSNR vs. SSIM},
  author={Hore, Alain and Ziou, Djemel},
  booktitle={2010 20th international conference on pattern recognition},
  pages={2366--2369},
  year={2010},
  organization={IEEE}
}

@inproceedings{panda2023agronav,
  title={Agronav: Autonomous navigation framework for agricultural robots and vehicles using semantic segmentation and semantic line detection},
  author={Panda, Shivam K and Lee, Yongkyu and Jawed, M Khalid},
  booktitle={2023 IEEE/CVF Conference on Computer Vision and Pattern Recognition Workshops (CVPRW)},
  pages={6272--6281},
  year={2023},
  organization={IEEE}
}

@article{zhu2024crops3d,
  title={Crops3D: a diverse 3D crop dataset for realistic perception and segmentation toward agricultural applications},
  author={Zhu, Jianzhong and Zhai, Ruifang and Ren, He and Xie, Kai and Du, Aobo and He, Xinwei and Cui, Chenxi and Wang, Yinghua and Ye, Junli and Wang, Jiashi and others},
  journal={Scientific Data},
  volume={11},
  number={1},
  pages={1438},
  year={2024},
  publisher={Nature Publishing Group UK, London}
}

@article{kheradmand20243d,
  title={3d gaussian splatting as markov chain monte carlo},
  author={Kheradmand, Shakiba and Rebain, Daniel and Sharma, Gopal and Sun, Weiwei and Tseng, Yang-Che and Isack, Hossam and Kar, Abhishek and Tagliasacchi, Andrea and Yi, Kwang Moo},
  journal={Advances in Neural Information Processing Systems},
  volume={37},
  pages={80965--80986},
  year={2024}
}

@article{kerbl20233d,
  title={3d gaussian splatting for real-time radiance field rendering.},
  author={Kerbl, Bernhard and Kopanas, Georgios and Leimk{\"u}hler, Thomas and Drettakis, George and others},
  journal={ACM Trans. Graph.},
  volume={42},
  number={4},
  pages={139--1},
  year={2023}
}

@inproceedings{espejel2024comparative,
  title={Comparative analysis of unity and gazebo simulators for digital twins of robotic tomato harvesting scenarios},
  author={Espejel Flores, Juan Pablo and Yilmaz, Abdurrahman and Soriano Avenda{\~n}o, Luis Arturo and Cielniak, Grzegorz},
  booktitle={Annual Conference Towards Autonomous Robotic Systems},
  pages={15--27},
  year={2024},
  organization={Springer}
}

@inproceedings{malik2024digital,
  title={A Digital Twin-Enabled Approach for Precision Weed Management in Specialty Crops using a 4-DoF Robotic System},
  author={Malik, Muneeb Elahi and Mahmud, Md Sultan},
  booktitle={2024 ASABE Annual International Meeting},
  pages={1},
  year={2024},
  organization={American Society of Agricultural and Biological Engineers}
}

@article{peng2025uavo,
  title={UAVO-NeRF: 3D reconstruction of orchards and semantic segmentation of fruit trees based on neural radiance field in UAV images},
  author={Peng, Hongxing and Guo, Shangkun and Zou, Xiangjun and Wang, Hongjun and Xiong, Juntao and Liang, Qijun},
  journal={Computers and Electronics in Agriculture},
  volume={237},
  pages={110631},
  year={2025},
  publisher={Elsevier}
}

@article{hani2020minneapple,
  title={MinneApple: a benchmark dataset for apple detection and segmentation},
  author={H{\"a}ni, Nicolai and Roy, Pravakar and Isler, Volkan},
  journal={IEEE Robotics and Automation Letters},
  volume={5},
  number={2},
  pages={852--858},
  year={2020},
  publisher={IEEE}
}

@article{kierdorf2023growliflower,
  title={GrowliFlower: An image time-series dataset for GROWth analysis of cauLIFLOWER},
  author={Kierdorf, Jana and Junker-Frohn, Laura Verena and Delaney, Mike and Olave, Mariele Donoso and Burkart, Andreas and Jaenicke, Hannah and Muller, Onno and Rascher, Uwe and Roscher, Ribana},
  journal={Journal of Field Robotics},
  volume={40},
  number={2},
  pages={173--192},
  year={2023},
  publisher={Wiley Online Library}
}

@article{de2022deep,
  title={Deep learning-based crop row detection for infield navigation of agri-robots},
  author={De Silva, Rajitha and Cielniak, Grzegorz and Wang, Gang and Gao, Junfeng},
  journal={Journal of field robotics},
  volume={41},
  number={7},
  pages={2299--2321},
  year={2024},
  publisher={Wiley Online Library}
}

@article{polvara2024bacchus,
  title={Bacchus Long-Term (BLT) data set: Acquisition of the agricultural multimodal BLT data set with automated robot deployment},
  author={Polvara, Riccardo and Molina, Sergi and Hroob, Ibrahim and Papadimitriou, Alexios and Tsiolis, Konstantinos and Giakoumis, Dimitrios and Likothanassis, Spiridon and Tzovaras, Dimitrios and Cielniak, Grzegorz and Hanheide, Marc},
  journal={Journal of Field Robotics},
  volume={41},
  number={7},
  pages={2280--2298},
  year={2024},
  publisher={Wiley Online Library}
}

@article{pire2019rosario,
  title={The Rosario dataset: Multisensor data for localization and mapping in agricultural environments},
  author={Pire, Taih{\'u} and Mujica, Mart{\'\i}n and Civera, Javier and Kofman, Ernesto},
  journal={The International Journal of Robotics Research},
  volume={38},
  number={6},
  pages={633--641},
  year={2019},
  publisher={SAGE Publications, Sage UK: London, England}
}

@inproceedings{barron2023zipnerf,
  title={Zip-nerf: Anti-aliased grid-based neural radiance fields},
  author={Barron, Jonathan T and Mildenhall, Ben and Verbin, Dor and Srinivasan, Pratul P and Hedman, Peter},
  booktitle={2023 IEEE/CVF International Conference on Computer Vision (ICCV)},
  pages={19640--19648},
  year={2023},
  organization={IEEE}
}

@article{xue2014gradient,
  title={Gradient magnitude similarity deviation: A highly efficient perceptual image quality index},
  author={Xue, Wufeng and Zhang, Lei and Mou, Xuanqin and Bovik, Alan C},
  journal={IEEE transactions on image processing},
  volume={23},
  number={2},
  pages={684--695},
  year={2013},
  publisher={IEEE}
}

@article{canny1986computational,
  title={A computational approach to edge detection},
  author={Canny, John},
  journal={IEEE Transactions on pattern analysis and machine intelligence},
  number={6},
  pages={679--698},
  year={1986},
  publisher={Ieee}
}

@article{arbelaez2011contour,
  title={Contour detection and hierarchical image segmentation},
  author={Arbelaez, Pablo and Maire, Michael and Fowlkes, Charless and Malik, Jitendra},
  journal={IEEE transactions on pattern analysis and machine intelligence},
  volume={33},
  number={5},
  pages={898--916},
  year={2010},
  publisher={IEEE}
}

@inproceedings{wu20253dgut,
  title={3DGUT: Enabling Distorted Cameras and Secondary Rays in Gaussian Splatting},
  author={Wu, Qi and Esturo, Janick Martinez and Mirzaei, Ashkan and Mo\"enne-Loccoz, Nicolas and Gojcic, Zan},
  booktitle={Proceedings of the IEEE/CVF Conference on Computer Vision and Pattern Recognition (CVPR)},
  month={June},
  year={2025},
  pages={26036-26046}
}

@article{3dgrt2024,
  title={3d gaussian ray tracing: Fast tracing of particle scenes},
  author={Moenne-Loccoz, Nicolas and Mirzaei, Ashkan and Perel, Or and De Lutio, Riccardo and Martinez Esturo, Janick and State, Gavriel and Fidler, Sanja and Sharp, Nicholas and Gojcic, Zan},
  journal={ACM Transactions on Graphics (TOG)},
  volume={43},
  number={6},
  pages={1--19},
  year={2024},
  publisher={ACM, New York, NY, USA}
}

@inproceedings{zhang2024GS-W,
  title={Gaussian in the wild: 3d gaussian splatting for unconstrained image collections},
  author={Zhang, Dongbin and Wang, Chuming and Wang, Weitao and Li, Peihao and Qin, Minghan and Wang, Haoqian},
  booktitle={European Conference on Computer Vision},
  pages={341--359},
  year={2024},
  organization={Springer}
}

@inproceedings{kulhanek2023tetranerf,
  title={Tetra-nerf: Representing neural radiance fields using tetrahedra},
  author={Kulhanek, Jonas and Sattler, Torsten},
  booktitle={2023 IEEE/CVF International Conference on Computer Vision (ICCV)},
  pages={18412--18423},
  year={2023},
  organization={IEEE}
}

@inproceedings{Taming3DGS,
  title={Taming 3dgs: High-quality radiance fields with limited resources},
  author={Mallick, Saswat Subhajyoti and Goel, Rahul and Kerbl, Bernhard and Steinberger, Markus and Carrasco, Francisco Vicente and De La Torre, Fernando},
  booktitle={SIGGRAPH Asia 2024 Conference Papers},
  pages={1--11},
  year={2024}
}

@article{chen2024pgsr,
  title={Pgsr: Planar-based gaussian splatting for efficient and high-fidelity surface reconstruction},
  author={Chen, Danpeng and Li, Hai and Ye, Weicai and Wang, Yifan and Xie, Weijian and Zhai, Shangjin and Wang, Nan and Liu, Haomin and Bao, Hujun and Zhang, Guofeng},
  journal={IEEE Transactions on Visualization and Computer Graphics},
  volume={31},
  number={9},
  pages={6100--6111},
  year={2024},
  publisher={IEEE}
}

@article{ye2025gsplat,
  title={gsplat: An open-source library for Gaussian splatting},
  author={Ye, Vickie and Li, Ruilong and Kerr, Justin and Turkulainen, Matias and Yi, Brent and Pan, Zhuoyang and Seiskari, Otto and Ye, Jianbo and Hu, Jeffrey and Tancik, Matthew and others},
  journal={Journal of Machine Learning Research},
  volume={26},
  number={34},
  pages={1--17},
  year={2025}
}

@inproceedings{scaffoldgs,
  title={Scaffold-gs: Structured 3d gaussians for view-adaptive rendering},
  author={Lu, Tao and Yu, Mulin and Xu, Linning and Xiangli, Yuanbo and Wang, Limin and Lin, Dahua and Dai, Bo},
  booktitle={2024 IEEE/CVF Conference on Computer Vision and Pattern Recognition (CVPR)},
  pages={20654--20664},
  year={2024},
  organization={IEEE}
}

@article{ren2024octree,
  title={Octree-gs: Towards consistent real-time rendering with lod-structured 3d gaussians},
  author={Ren, Kerui and Jiang, Lihan and Lu, Tao and Yu, Mulin and Xu, Linning and Ni, Zhangkai and Dai, Bo},
  journal={arXiv preprint arXiv:2403.17898},
  year={2024}
}

@inproceedings{nerfstudio,
  title={Nerfstudio: A modular framework for neural radiance field development},
  author={Tancik, Matthew and Weber, Ethan and Ng, Evonne and Li, Ruilong and Yi, Brent and Wang, Terrance and Kristoffersen, Alexander and Austin, Jake and Salahi, Kamyar and Ahuja, Abhik and others},
  booktitle={ACM SIGGRAPH 2023 conference proceedings},
  pages={1--12},
  year={2023}
}

@article{hierarchicalgaussians24,
  title={A hierarchical 3d gaussian representation for real-time rendering of very large datasets},
  author={Kerbl, Bernhard and Meuleman, Andreas and Kopanas, Georgios and Wimmer, Michael and Lanvin, Alexandre and Drettakis, George},
  journal={ACM Transactions On Graphics (TOG)},
  volume={43},
  number={4},
  pages={1--15},
  year={2024},
  publisher={ACM, New York, NY, USA}
}

@inproceedings{wang2025vggt,
  title={Vggt: Visual geometry grounded transformer},
  author={Wang, Jianyuan and Chen, Minghao and Karaev, Nikita and Vedaldi, Andrea and Rupprecht, Christian and Novotny, David},
  booktitle={2025 IEEE/CVF Conference on Computer Vision and Pattern Recognition (CVPR)},
  pages={5294--5306},
  year={2025},
  organization={IEEE}
}

@inproceedings{schoenberger2016sfm,
  title={Structure-from-motion revisited},
  author={Schonberger, Johannes L and Frahm, Jan-Michael},
  booktitle={Proceedings of the IEEE conference on computer vision and pattern recognition},
  pages={4104--4113},
  year={2016}
}

@article{kulhanek2025nerfbaselines,
  title={Nerfbaselines: Consistent and reproducible evaluation of novel view synthesis methods},
  author={Kulhanek, Jonas and Sattler, Torsten},
  journal={Advances in Neural Information Processing Systems},
  volume={38},
  year={2026}
}

@inproceedings{levy2023seathru,
  title={Seathru-nerf: Neural radiance fields in scattering media},
  author={Levy, Deborah and Peleg, Amit and Pearl, Naama and Rosenbaum, Dan and Akkaynak, Derya and Korman, Simon and Treibitz, Tali},
  booktitle={2023 IEEE/CVF Conference on Computer Vision and Pattern Recognition (CVPR)},
  pages={56--65},
  year={2023},
  organization={IEEE}
}

@article{li2025survey,
  title={A survey on 3D reconstruction techniques in plant phenotyping: From classical methods to Neural Radiance Fields (NeRF), 3D Gaussian Splatting (3DGS), and beyond},
  author={Li, Jiajia and Qi, Xinda and Nabaei, Seyed Hamidreza and Liu, Meiqi and Chen, Dong and Sun, Qi and Zhang, Xin and Yin, Xunyuan and Li, Zhaojian},
  journal={Plant Phenomics},
  pages={100137},
  year={2025},
  publisher={Elsevier}
}

@inproceedings{zhang2025wheat3dgs,
  title={Wheat3dgs: In-field 3d reconstruction, instance segmentation and phenotyping of wheat heads with gaussian splatting},
  author={Zhang, Daiwei and Gajardo, Joaquin and Medic, Tomislav and Katircioglu, Isinsu and Boss, Mike and Kirchgessner, Norbert and Walter, Achim and Roth, Lukas},
  booktitle={2025 IEEE/CVF Conference on Computer Vision and Pattern Recognition Workshops (CVPRW)},
  pages={5360--5370},
  year={2025},
  organization={IEEE}
}

@article{akhtar2024unlocking,
  title={Unlocking plant secrets: A systematic review of 3D imaging in plant phenotyping techniques},
  author={Akhtar, Muhammad Salman and Zafar, Zuhair and Nawaz, Raheel and Fraz, Muhammad Moazam},
  journal={Computers and Electronics in Agriculture},
  volume={222},
  pages={109033},
  year={2024},
  publisher={Elsevier}
}

@article{an2017quantifying,
  title={Quantifying time-series of leaf morphology using 2D and 3D photogrammetry methods for high-throughput plant phenotyping},
  author={An, Nan and Welch, Stephen M and Markelz, RJ Cody and Baker, Robert L and Palmer, Christine M and Ta, James and Maloof, Julin N and Weinig, Cynthia},
  journal={Computers and Electronics in Agriculture},
  volume={135},
  pages={222--232},
  year={2017},
  publisher={Elsevier}
}

@article{agrigaussian2026,
  title={AgriGaussian: A low-cost 3D reconstruction method for high-fidelity plant architecture analysis},
  author={An, Qilin and Li, Guangrui and Deng, Xue and Cui, Yongzhi and Guo, Quanfeng and Liu, Yangchun and Zhou, Liming and Wei, Liguo and Zhao, Bo},
  journal={Artificial Intelligence in Agriculture},
  year={2026},
  publisher={Elsevier}
}

@inproceedings{chang2026nirsplat,
  title={Reconstruction Using the Invisible: Intuition from NIR and Metadata for Enhanced 3D Gaussian Splatting},
  author={Chang, Gyusam and Vu, Tuan-Anh and Alumootil, Vivek and Song, Harris and Pham, Deanna and Kim, Sangpil and Jawed, M Khalid},
  booktitle={Proceedings of the AAAI Conference on Artificial Intelligence},
  volume={40},
  number={4},
  pages={2707--2715},
  year={2026}
}

@article{cuaran2023undercanopy,
  title={Under-canopy dataset for advancing simultaneous localization and mapping in agricultural robotics},
  author={Cuaran, Jose and Baquero Velasquez, Andres Eduardo and Valverde Gasparino, Mateus and Uppalapati, Naveen Kumar and Sivakumar, Arun Narenthiran and Wasserman, Justin and Huzaifa, Muhammad and Adve, Sarita and Chowdhary, Girish},
  journal={The International Journal of Robotics Research},
  volume={43},
  number={6},
  pages={739--749},
  year={2024},
  publisher={SAGE Publications, Sage UK: London, England}
}

@inproceedings{marks2024bonnbeetclouds,
  title={BonnBeetClouds3D: A dataset towards point cloud-based organ-level phenotyping of sugar beet plants under real field conditions},
  author={Marks, Elias and B{\"o}mer, Jonas and Magistri, Federico and Sag, Anurag and Behley, Jens and Stachniss, Cyrill},
  booktitle={2024 IEEE/RSJ International Conference on Intelligent Robots and Systems (IROS)},
  pages={1804--1811},
  year={2024},
  organization={IEEE}
}

@article{subeesh2025digitaltwin,
  title={Agricultural digital twin for smart farming: A review},
  author={Subeesh, A and Chauhan, Naveen},
  journal={Green Technologies and Sustainability},
  pages={100299},
  year={2025},
  publisher={Elsevier}
}

@article{wu2026gc3dgspheno,
  title={GC-3DGS-pheno: geometry-constrained 3d Gaussian splatting for affordable plant point cloud reconstruction and high-precision phenotyping},
  author={Wu, Yandong and Wen, Weiliang and Zhu, Yongji and Wu, Zhangbin and Li, Yinglun and Guo, Xinyu and Zhao, Chunjiang},
  journal={Computers and Electronics in Agriculture},
  pages={112467},
  year={2026},
  publisher={Elsevier}
}

@article{chen2025gsag,
  title={High-fidelity 3D reconstruction of peach orchards using a 3DGS-Ag model},
  author={Chen, Yanan and Xiao, Ke and Gao, Guandong and Zhang, Fan},
  journal={Computers and Electronics in Agriculture},
  volume={234},
  pages={110225},
  year={2025},
  publisher={Elsevier}
}

@article{shen2025plantgaussian,
  title={PlantGaussian: Exploring 3D Gaussian splatting for cross-time, cross-scene, and realistic 3D plant visualization and beyond},
  author={Shen, Peng and Jing, Xueyao and Deng, Wenzhe and Jia, Hanyue and Wu, Tingting},
  journal={The Crop Journal},
  volume={13},
  number={2},
  pages={607--618},
  year={2025},
  publisher={Elsevier}
}

@article{wang2025p3dfusion,
  title={P3DFusion: A cross-scene and high-fidelity 3D plant reconstruction framework empowered by vision foundation models and 3D Gaussian splatting},
  author={Wang, Zhi and Xiao, Shunfu and Miao, Zhuang and Liu, Ruixue and Chen, Haochong and Wang, Qing and Shao, Ke and Wang, Ruili and Ma, Yuntao},
  journal={European Journal of Agronomy},
  volume={171},
  pages={127811},
  year={2025},
  publisher={Elsevier}
}

@article{jiang2025cotton3dgaussians,
  title={Cotton3DGaussians: Multiview 3D Gaussian Splatting for boll mapping and plant architecture analysis},
  author={Jiang, Lizhi and Sun, Jin and Chee, Peng W and Li, Changying and Fu, Longsheng},
  journal={Computers and Electronics in Agriculture},
  volume={234},
  pages={110293},
  year={2025},
  publisher={Elsevier}
}

@article{shen2026thingaussian,
  title={Thin Gaussian Splatting for high-quality geometric reconstruction of peach tree},
  author={Shen, Bowei and Xiao, Ke and Zhao, Jin and Gao, Qiyuan},
  journal={Computers and Electronics in Agriculture},
  volume={254},
  pages={112232},
  year={2026},
  publisher={Elsevier}
}

@article{zuniga2019multisensor,
  title={Automatic multi-sensor extrinsic calibration for mobile robots},
  author={Zu{\~n}iga-No{\"e}l, David and Ruiz-Sarmiento, Jose-Raul and Gomez-Ojeda, Ruben and Gonzalez-Jimenez, Javier},
  journal={IEEE Robotics and Automation Letters},
  volume={4},
  number={3},
  pages={2862--2869},
  year={2019},
  publisher={IEEE}
}

@inproceedings{carion2026sam3,
  title={Sam 3: Segment anything with concepts},
  author={Carion, Nicolas and Gustafson, Laura and Hu, Yuan-Ting and Debnath, Shoubhik and Hu, Ronghang and Suris Coll-Vinent, Didac and Ryali, Chaitanya and Alwala, Kalyan Vasudev and Khedr, Haitham and Huang, Andrew and others},
  booktitle={International conference on learning representations},
  volume={2026},
  pages={138846--138923},
  year={2026}
}

@inproceedings{ye2024absgs,
  title={Absgs: Recovering fine details in 3d gaussian splatting},
  author={Ye, Zongxin and Li, Wenyu and Liu, Sidun and Qiao, Peng and Dou, Yong},
  booktitle={Proceedings of the 32nd ACM international conference on multimedia},
  pages={1053--1061},
  year={2024}
}

@inproceedings{girish2024eagles,
  title={Eagles: Efficient accelerated 3d gaussians with lightweight encodings},
  author={Girish, Sharath and Gupta, Kamal and Shrivastava, Abhinav},
  booktitle={European Conference on Computer Vision},
  pages={54--71},
  year={2024},
  organization={Springer}
}

@inproceedings{zhang2024fregs,
  title={Fregs: 3d gaussian splatting with progressive frequency regularization},
  author={Zhang, Jiahui and Zhan, Fangneng and Xu, Muyu and Lu, Shijian and Xing, Eric},
  booktitle={2024 IEEE/CVF Conference on Computer Vision and Pattern Recognition (CVPR)},
  pages={21424--21433},
  year={2024},
  organization={IEEE}
}

@article{li2026hf,
  title={HF-GS: Plug-and-play high-frequency-guided Gaussian allocation for 3D Gaussian splatting},
  author={Li, Zhanrong and He, Longning and Nong, Yijia and He, Junle},
  journal={Computers \& Graphics},
  pages={104606},
  year={2026},
  publisher={Elsevier}
}
